\documentclass[11pt]{article}

\usepackage[preprint]{acl}
\usepackage[T1]{fontenc}
\usepackage{times}
\usepackage{latexsym}
\usepackage{amsmath,amssymb}
\usepackage{booktabs}
\usepackage{graphicx}
\usepackage{placeins}
\usepackage{microtype}
\usepackage{multirow}
\usepackage{array}
\usepackage{tabularx}
\usepackage{xcolor}
\usepackage{colortbl}
\usepackage{tikz}
\usetikzlibrary{arrows.meta,positioning}

\graphicspath{{figures/}}
\newcommand{\gint}{\gamma}
\newcommand{\dmse}{\Delta\mathrm{MSE}}

\newcolumntype{Y}{>{\raggedright\arraybackslash}X}
\definecolor{cueblue}{HTML}{236589}
\definecolor{cueorange}{HTML}{B86432}
\definecolor{cueteal}{HTML}{247A71}
\definecolor{tablegray}{gray}{0.94}

\title{Beyond Correct Answers:\\
How Compositional Cues Shape Visual Judgments}
\author{%
Xiaoyu Ma \quad Zheyuan Zhang \quad Kaiwen Shi \quad Yanfang Ye$^{\dagger}$ \\
{\normalfont University of Notre Dame} \\
{\normalfont\small\texttt{\{xma6,zzhang42,kshi3,yye7\}@nd.edu}} \\
{\normalfont\small $\dagger$Corresponding author.}%
}

\begin{document}
\maketitle

\begin{abstract}
Visual question answering requires models to assign attributes and
relations to the correct objects, a problem known as visual binding.
Structured prompts make these assignments explicit through labeled fields.
Yet an improvement from the full prompt does not reveal which labels
help or how they interact.
First, we isolate field-label effects across ten models and six tasks
by holding the image and claim fixed. These comparisons reveal
differences that overall accuracy can hide. With a relation label
present, adding object-role labels helps Qwen3 but hurts Qwen2 on Spatial.
Second, we compare auxiliary hints that preserve or swap the question's
role assignments against a no-hint baseline. Across two image sets,
Qwen3's advantage with preserving hints mainly reflects errors caused
by swapped hints. Preserving hints remain near baseline. Field order
further affects the harm. Third, we test clearer hint wording and direct
selection of the relation shown in the image. Clearer hints improve
text matching but do not consistently improve visual judgments.
In contrast, relation selection partly reduces errors for some hints
in Qwen3 and GPT-4.1-mini.
\end{abstract}
\section{Introduction}
\label{sec:intro}

Vision--language models (VLMs) rely on object recognition to describe
visual scenes and answer questions about them
\citep{li2023blip2,liu2023visual}.
But recognizing the objects alone is not enough.
The model must also determine which object has an attribute or occupies
a relation role.
For example, a model may recognize both dogs yet confuse which is left
of the other.
Assigning attributes and roles to objects is known as visual binding
\citep{greff2020binding,campbell2025binding}.

\begin{figure}[t]
\centering
\includegraphics[width=\columnwidth]{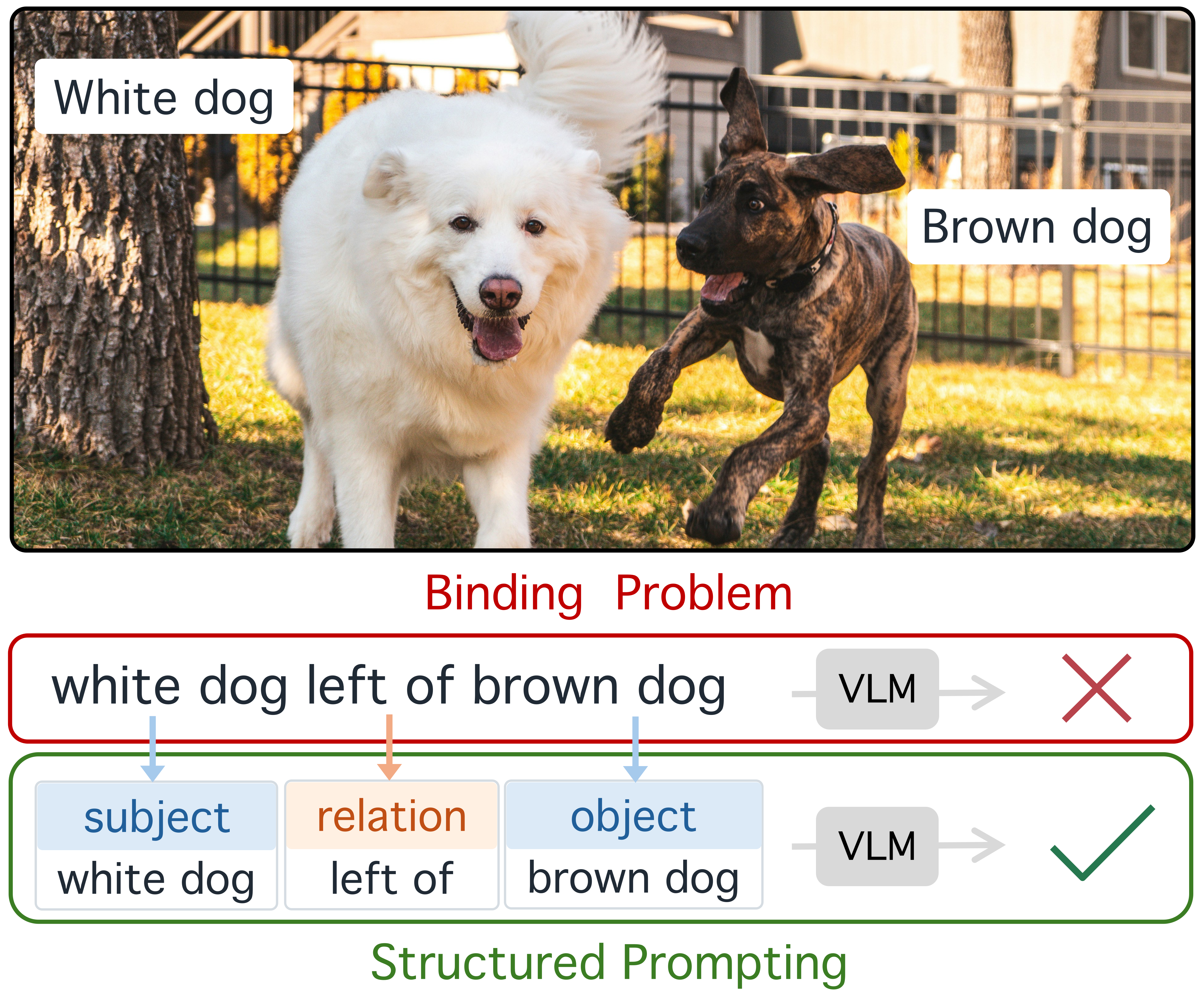}
\caption{Motivating schematic, not an observed answer change.
Photo: \href{https://unsplash.com/photos/two-dogs-running-in-the-grass-near-a-tree-P7ngl_WZyKI}{Caleb Bennetts/Unsplash}.}
\label{fig:role-prompts}
\end{figure}

Compositional evaluations test this distinction by contrasting
descriptions that share scene elements but assign
them differently \citep{thrush2022winoground,yuksekgonul2023aro}.
To support these judgments, structured prompting represents objects,
attributes, and relations explicitly.
Scene-graph prompting and concept-based reasoning use these representations
to improve judgments without retraining \citep{mitra2024ccot,sinha2025cocotree}.
However, these approaches can jointly change content, organization,
and reasoning.
These overall gains do not isolate the effects of labels such as
\texttt{subject} and \texttt{relation}.
Figure~\ref{fig:role-prompts} illustrates this motivation: explicit role
labels could help resolve a binding error while preserving both the
image and the claim being judged.

The contribution of a label may depend on the other labels present.
For example, object-role labels may become useful only when the relation
is also labeled.
The error type may also matter.
A claim can name the wrong object or state the wrong attribute value.
Aggregate accuracy alone may not reveal which components help.
\textbf{RQ1. How do explicit field labels affect visual judgments,
separately and together?}

Beyond labels within a claim, role assignments can be supplied as
auxiliary hints beside an unchanged question.
Structured reasoning can reuse model-generated records as context
\citep{mitra2024ccot}. Errors in these records can mislead later visual
answers \citep{zhong2024snowball,deng2025words}.
Such hints may preserve the question's assignment or swap its
participants. Yet a preserving hint adds no new visual evidence about
the claim's truth.
A preserving-over-swapped advantage may therefore reflect useful
guidance, errors introduced by the swap, or both.
A no-hint baseline distinguishes these effects.
Prompt presentation can also affect responses \citep{sclar2024formatspread}.
Field-order controls separate this influence from the role assignment.
\textbf{RQ2. Does a role-preserving advantage reflect genuine help
or harm from swapped hints?}

Diagnosing hint-induced errors also motivates interventions.
Clearer wording may improve role interpretation.
Even so, the model must judge whether the image supports the claim.
If errors persist despite correct interpretation, changing the judgment
task offers another possibility.
Direct relation selection asks what the image depicts instead of
whether a supplied claim is true.
\textbf{RQ3. Can hint rewriting or task reformulation reduce
hint-induced errors without retraining?}

Across ten models and six tasks, our fixed-content label comparisons
reveal differences hidden by overall accuracy.
On Spatial, adding object-role labels alongside a relation label helps
Qwen3 but hurts Qwen2.
For auxiliary hints, matched comparisons across two image sets show that
Qwen3's preserving-over-swapped advantage mainly reflects swapped-hint harm.
Preserving hints remain near the no-hint baseline, while canonical field
order amplifies the harm.
Guided by this diagnosis, we test hint rewrites and relation selection
in Qwen3 and GPT-4.1-mini (mini).
Clearer wording improves text matching but does not consistently improve
visual judgments.
By contrast, relation selection partly reduces errors for some hints.
These contrasts account for the no-hint task difference.

Our contributions are threefold.
\begin{enumerate}
\item \textbf{Controlled role-label decomposition.}
We introduce fixed-content comparisons that separate responses to
object and value errors and identify component effects and interactions.
\item \textbf{Benefit--harm diagnosis of role hints.}
We combine no-hint baselines, field-order controls, and paired-answer
analysis to distinguish useful guidance from added errors.
\item \textbf{Diagnosis-guided interventions.}
We compare hint rewriting with relation selection, adjust for no-hint
task differences, and count corrections and new errors.
\end{enumerate}

\section{Related Work}
\label{sec:related}

\paragraph{Visual Binding and Compositionality.}
Compositional evaluations distinguish recognizing scene elements from
recovering their assignments, with controls for linguistic shortcuts
\citep{thrush2022winoground,yuksekgonul2023aro,ma2024sade}.
Cross-modal entity linking also requires matching visual objects to
textual references \citep{alonso2025mate}. Structured prompting and
relation-guided reasoning support these assignments
\citep{mitra2024ccot,sinha2025cocotree,bukkapatnam2026reground}.
Scene-graph ablations test both content and JSON formatting
\citep{mitra2024ccot}. We hold claim content and field order fixed to
isolate which role labels drive changes in visual judgments,
both individually and in combination with other labels.

\begin{figure*}[t]
\centering
\includegraphics[width=\textwidth]{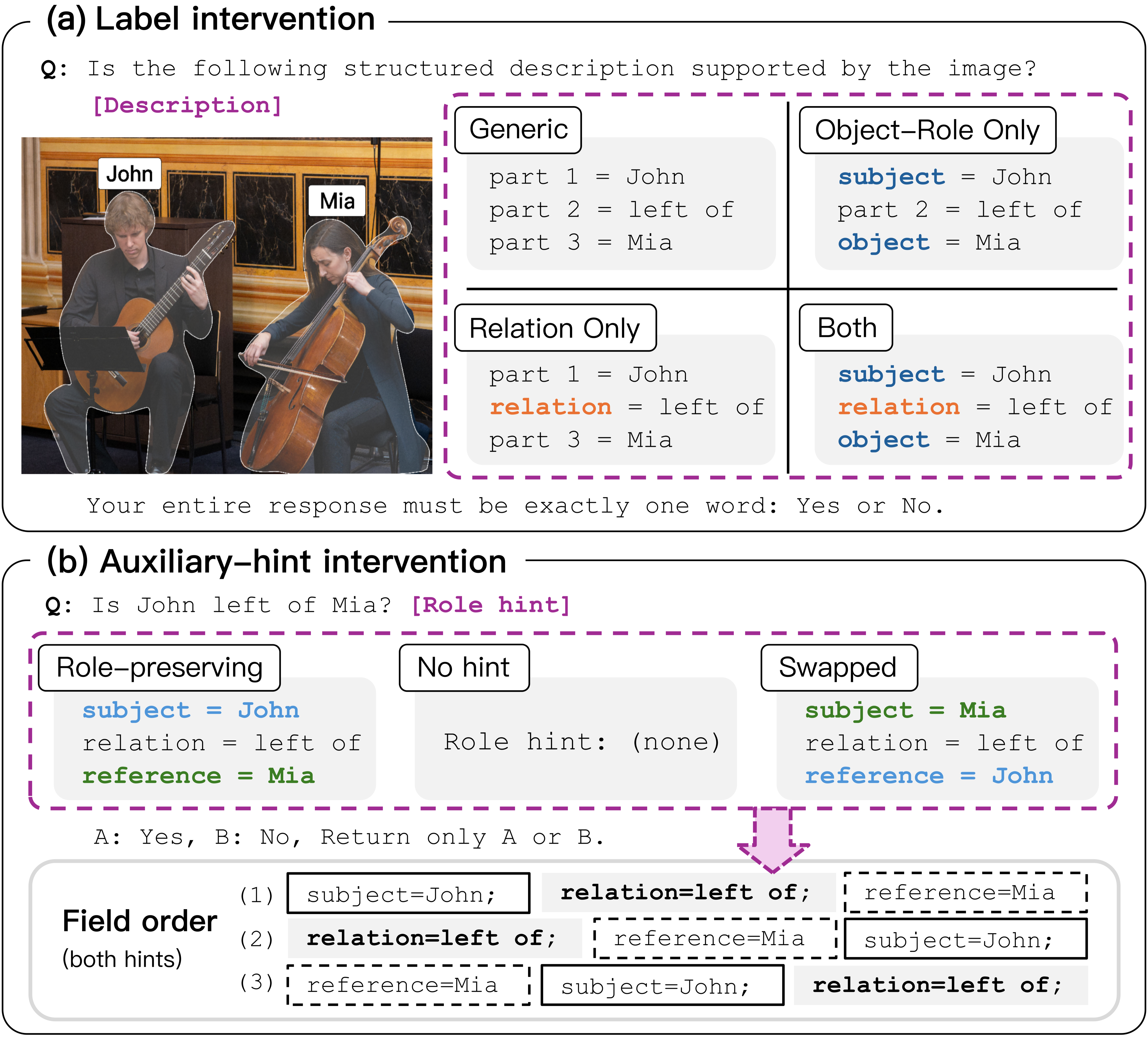}
\caption{Prompt excerpts, with spacing adjusted for display.
Each box represents a separate condition.
(a) Only object-role and relation labels vary.
(b) The question is fixed, and both hint types use all three field orders.
John/Mia are illustrative names, and A/B
are answer options. Annotated photo:
\href{https://commons.wikimedia.org/wiki/File:Cristina_Basili_Timotej_Kosovinc_2025_BHO-7257.jpg}{Bernhard Holub},
\href{https://creativecommons.org/licenses/by-sa/4.0/}{CC BY-SA 4.0}.}
\label{fig:cues}
\end{figure*}

\begin{table*}[t]
\centering
\begingroup
\footnotesize
\setlength{\tabcolsep}{2pt}
\renewcommand{\arraystretch}{1.20}
\newcommand{\tOneAcc}{Acc.}
\newcommand{\tOneGamma}[1]{$\gamma_{\mathrm{#1}}$}
\definecolor{gammaPositive}{HTML}{28699A}
\definecolor{gammaNegative}{HTML}{AE4542}
\newcommand{\tOnePos}[1]{\textcolor{gammaPositive}{$#1$}}
\newcommand{\tOneNeg}[1]{\textcolor{gammaNegative}{$#1$}}
\begin{tabular*}{\linewidth}{@{\extracolsep{\fill}}ll*{3}{r@{\hspace{3.5pt}}r}@{\hspace{9pt}}*{3}{r@{\hspace{3.5pt}}r}@{}}
\toprule
 & & \multicolumn{6}{c}{\textbf{Attributes: error-type contrast (NLL)}}
 & \multicolumn{6}{c}{\textbf{Relations: label interaction (log odds)}} \\
\cmidrule(lr){3-8}\cmidrule(l){9-14}
 & & \multicolumn{2}{c}{Material}
 & \multicolumn{2}{c}{Size}
 & \multicolumn{2}{c}{Color}
 & \multicolumn{2}{c}{Spatial}
 & \multicolumn{2}{c}{Holding}
 & \multicolumn{2}{c}{Attached-to} \\
\cmidrule(lr){3-4}\cmidrule(lr){5-6}\cmidrule(lr){7-8}
\cmidrule(lr){9-10}\cmidrule(lr){11-12}\cmidrule(l){13-14}
Model & Readout
 & \tOneAcc & \tOneGamma{attr} & \tOneAcc & \tOneGamma{attr} & \tOneAcc & \tOneGamma{attr}
 & \tOneAcc & \tOneGamma{rel} & \tOneAcc & \tOneGamma{rel} & \tOneAcc & \tOneGamma{rel} \\
\midrule
LLaVA & L1 & 88.3 & \tOnePos{+.038} & 75.8 & \tOneNeg{-.012} & 94.5 & \tOnePos{+.057} & 52.0 & \tOnePos{+.008} & 64.1 & \tOnePos{+.031} & 67.2 & \tOnePos{+.090} \\
Qwen2 & L1 & 93.8 & \tOneNeg{-.008} & 83.6 & \tOneNeg{-.144} & 96.1 & \tOneNeg{-.000} & 79.4 & \tOneNeg{-.142} & \textbf{84.4} & \tOnePos{+.104} & 82.8 & \tOnePos{+.080} \\
Qwen3 & L1 & 95.8 & \tOneNeg{-.078} & 83.1 & \tOneNeg{-.140} & 96.4 & \tOneNeg{-.072} & 81.9 & \tOneNeg{-.072} & 78.1 & \tOnePos{+.684} & \textbf{93.8} & \tOnePos{+1.135} \\
Gemma4 & L1 & 38.0 & \tOneNeg{-.013} & 42.7 & \tOnePos{+.037} & 46.1 & \tOneNeg{-.050} & 73.3 & \tOnePos{+.402} & 71.9 & \tOneNeg{-1.330} & 81.2 & \tOnePos{+.175} \\
InternVL3.5 & LS & 92.4 & \tOneNeg{-.105} & 77.6 & \tOneNeg{-.179} & 97.7 & \tOneNeg{-.134} & \textbf{84.4} & \tOnePos{+.101} & \textbf{84.4} & \tOnePos{+.248} & 81.2 & \tOnePos{+.152} \\
GLM-4.1V & LS & 95.3 & \tOnePos{+.011} & 79.2 & \tOnePos{+.001} & \textbf{97.9} & \tOnePos{+.011} & 63.2 & \tOneNeg{-.063} & 78.1 & \tOnePos{+.245} & 81.2 & \tOnePos{+.317} \\
Phi-4 & LS & 92.4 & \tOnePos{+.142} & \textbf{85.2} & \tOneNeg{-.458} & 95.3 & \tOnePos{+.074} & 83.7 & \tOnePos{+.161} & 79.7 & \tOnePos{+.012} & 78.1 & \tOnePos{+.008} \\
NVIDIA Llama & API & 91.7 & \tOnePos{+.099} & 79.2 & \tOnePos{+.032} & 94.0 & \tOnePos{+.121} & 60.5 & \tOneNeg{-.004} & 78.1 & \tOnePos{+.197} & 78.1 & \tOnePos{+.162} \\
GPT-4.1-mini & API & \textbf{96.1} & \tOneNeg{-.483} & 81.0 & \tOneNeg{-1.206} & 96.9 & \tOneNeg{-.745} & 76.8 & \tOneNeg{-.177} & 81.2 & \tOnePos{+1.039} & 87.5 & \tOnePos{+1.971} \\
GPT-4o & API & 95.6 & \tOneNeg{-.218} & 78.4 & \tOneNeg{-.741} & 96.1 & \tOneNeg{-.053} & 69.8 & \tOneNeg{-.376} & 81.2 & \tOneNeg{-.664} & 89.1 & \tOnePos{+.001} \\
\bottomrule
\end{tabular*}
\par
\endgroup
\caption{Ordinary accuracy and label effects across ten models and six tasks.
Bold marks the highest accuracy per task. Blue and red distinguish
positive and negative $\gamma$, not performance ranks.
Accuracy is in \% and score-based $\gamma$ in nats.
L1 and LS denote local first-token and full-answer scores, respectively.}
\label{tab:fullmatrix}
\end{table*}

\paragraph{Prompt Sensitivity and Textual Interference.}
Prompt studies quantify performance variation across formats and
response sensitivity across rephrasings
\citep{sclar2024formatspread,errica2025wrong, zhang2026mapro}.
In VLMs, language priors and conflicting text can override visual
evidence \citep{lee2025vlind,deng2025words,pezeshkpour2025mixed}.
Textual interference depends on the task, model, and input order.
Our focus is the source of a role-preserving advantage. Comparisons
with no hint distinguish genuine benefit from swapped-hint harm,
while order changes locate the influence of presentation. This
diagnosis motivates interventions on both the hint and the judgment task.

\paragraph{Consistency-Based Answer Verification.}
Variation across related questions supports visual-answer checking
\citep{khan2024consistency,yang2025sources,liu2026taco, zhang2025agentrouter}. We compare role-explicit
and ordinary checks at the same query budget as a reliability application.
Appendix~\ref{app:related} discusses connections to LLM uncertainty estimation.

\section{Preliminaries}
\label{sec:prelim}

\paragraph{Visual binding.}
A visual claim specifies not only scene elements but also their
assignment to roles. Let $\mathcal{O}_I$ be the object instances in image
$I$. A claim $c$ with $k$ object positions has a binding
\begin{equation}
b:\{1,\ldots,k\}\longrightarrow\mathcal{O}_I.
\label{eq:binding}
\end{equation}
An attribute claim has the form $a(b(1))=v$, where $a$ is an attribute
and $v$ its value. A relation claim has the form $r(b(1),b(2))$.
The assignment $b$ identifies which object has the proposed value or
occupies each relation role. The image determines whether the resulting
claim $c[b]$ is true.

\paragraph{Two error types.}
Starting from a true attribute claim, a wrong-object alternative changes
$b$ while retaining the attribute and value. A wrong-value alternative
changes $v$ while retaining the object and attribute. We retain an
alternative as a negative only when the change makes the claim false.

\paragraph{Labels, values, and order.}
In \texttt{subject=A}, \texttt{subject} is a field label and \texttt{A}
is its value. Renaming it \texttt{part1=A} removes the explicit role
label but preserves the value. Reordering whole fields preserves their
label--value assignments.

\section{Methodology}
\label{sec:method}

Our compositional cues are claim labels and auxiliary role hints
(Figure~\ref{fig:cues}). Matched interventions isolate label effects
and distinguish hint benefit from harm. Paired errors guide intervention.

\subsection{Role-Label Interventions}

\paragraph{Attribute comparisons.}
\label{sec:attribute}
For attributes, we compare explicit field names with generic names on
choices between a true claim and a wrong-object or wrong-value alternative.
Let $B$ and $F$ be the mean
changes in the correct answer's negative log-likelihood (NLL),
explicit minus generic, for these two error types. Their difference is
\begin{equation}
\gint_{\rm attr}=B-F.
\label{eq:gamma-attribute}
\end{equation}
A negative value means that labels reduce NLL more, or increase it less,
for wrong-object judgments. This contrast crosses label naming with
error type. We compute attribute NLL from pooled prompt margins
(Appendix~\ref{app:matrix}).

\paragraph{Relation comparisons.}
\label{sec:relation}

For relations, we represent a proposition with three fields:
\texttt{subject=S}, \texttt{relation=R}, and \texttt{object=O}.
The two participant labels form the object-role component.
Replacing either component with generic names gives four conditions:
Generic, Object-role only, Relation only, and Both.
Participants, relations, and field order remain fixed.

Let $z_{xy}$ denote support for the correct answer, where $x$ indicates
explicit object-role labels and $y$ the relation label. We use
$z=\log(p_{\rm correct}/p_{\rm incorrect})$, summing accepted variants
of each answer label. Positive values favor the correct answer.
The interaction compares the effect of adding object-role labels with
and without a relation label:
\begin{equation}
\gint_{\rm rel}=(z_{11}-z_{01})-(z_{10}-z_{00}).
\label{eq:gamma}
\end{equation}
The interaction measures how one label changes the effect of the other.
The net change from using both labels is $z_{11}-z_{00}$.
We report condition accuracies alongside these contrasts to distinguish
label interaction from overall improvement.

\begin{table*}[t]
\centering
\begingroup
\small
\setlength{\tabcolsep}{2.5pt}
\renewcommand{\arraystretch}{1.20}
\begin{tabular*}{\linewidth}{@{\extracolsep{\fill}}lrrrrr@{\hspace{2pt}}lr@{\hspace{2pt}}l@{}}
\toprule
 & \multicolumn{4}{c}{Condition accuracy (\%)}
 & \multicolumn{4}{c}{Accuracy change (pp)} \\
\cmidrule(lr){2-5}\cmidrule(l){6-9}
Model & Generic & Object-role & Relation & Both
 & \multicolumn{2}{c}{Both $-$ Generic} & \multicolumn{2}{c}{Both $-$ Relation} \\
\midrule
LLaVA & \textbf{50.26} & 50.13 & 50.13 & 49.74
 & $-0.52$ & $[-1.30,0.26]$ & $-0.39$ & $[-1.04,0.26]$ \\
Qwen2 & 55.86 & 63.54 & \textbf{67.06} & 63.67
 & $+7.81$ & $[3.78,11.72]$ & \cellcolor{tablegray}$-3.39$ & \cellcolor{tablegray}$[-6.12,-0.65]$ \\
Qwen3 & 59.77 & 68.36 & 68.36 & \textbf{73.05}
 & $+13.28$ & $[9.51,16.93]$ & \cellcolor{tablegray}$+4.69$ & \cellcolor{tablegray}$[2.21,7.16]$ \\
Gemma4 & 49.48 & 50.65 & 58.07 & \textbf{58.59}
 & $+9.11$ & $[5.60,12.63]$ & $+0.52$ & $[-2.47,3.52]$ \\
GPT-4.1-mini & 63.67 & 64.06 & 63.15 & \textbf{64.97}
 & $+1.30$ & $[-1.43,4.04]$ & $+1.82$ & $[-0.52,4.30]$ \\
NVIDIA Llama & 51.82 & 51.56 & 53.13 & \textbf{55.86}
 & $+4.04$ & $[1.95,6.25]$ & $+2.73$ & $[0.26,5.34]$ \\
\bottomrule
\end{tabular*}
\par
\endgroup
\caption{Spatial accuracy with images, claim values, field order, and
Yes/No responses fixed. Bold marks each model's highest accuracy.
Shading highlights opposite effects of adding object-role labels to
Relation only. Brackets give paired 95\% intervals over 48 image pairs.
Ordinary-question accuracy (Table~\ref{tab:fullmatrix}) uses a separate protocol.}
\label{tab:cue-ablation}
\end{table*}

\subsection{Auxiliary-Hint Interventions}
\label{sec:hint-design}

Here the original question is unchanged.
We construct each hint directly from the question.
The role-preserving hint ($C$) retains its role assignments.
The swapped hint ($S$) exchanges only the two participant values,
leaving the relation unchanged.
These names describe the question's roles, not image truth.
The model is instructed to answer the question even if the hint is
inaccurate. Each hint contains the same
three fields: \texttt{subject}, \texttt{relation}, and \texttt{reference}.
We present them in canonical order ($K$) or one of two cyclic rotations.
The baseline has no hint ($N$).

Let $A_{CK},A_{CR},A_{SK},A_{SR}$ be the four accuracies, where $R$
equally averages the two rotations. We give canonical and mean rotated
order equal weight, so $A_C=(A_{CK}+A_{CR})/2$ and
$A_S=(A_{SK}+A_{SR})/2$. The preserving-over-swapped gap decomposes as
\begin{align}
H &= A_C-A_N, \qquad J=A_N-A_S,\\
D_C &= A_C-A_S=H+J.
\label{eq:help-harm}
\end{align}
Here $H$ is the preserving-hint benefit and $J$ the swapped-hint loss.
To make the direction consistent, our displays show $H$ and $-J$:
positive values always mean an increase over no hint. Reporting both
terms reveals the source of the relative gap.

The content-by-order interaction is
\begin{equation}
I=(A_{CK}-A_{SK})-(A_{CR}-A_{SR}).
\label{eq:gamma-accuracy}
\end{equation}
It asks whether the role gap changes with field order. Unlike
Eq.~\ref{eq:gamma}, its factors are hint content and order, and its
outcome is accuracy. All accuracy contrasts are reported in percentage
points (pp).

We also pair answers to opposing claims about the same image.
The four patterns are both correct, both reversed, both No, and
both Yes. They distinguish rejection of true claims from a reversal
of both judgments.

\begin{table*}[t]
\centering
\small
\setlength{\tabcolsep}{4pt}
\newcommand{\mci}[3]{$#1$\;$[#2,#3]$}
\begin{tabularx}{\linewidth}{Xrrrr}
\toprule
 & \multicolumn{1}{c}{Accuracy (\%)} & \multicolumn{1}{c}{Hint contrast (pp)}
 & \multicolumn{2}{c}{\textbf{Change from no hint (pp)}} \\
\cmidrule(lr){2-2}\cmidrule(lr){3-3}\cmidrule(l){4-5}
Model & \multicolumn{1}{c}{No hint} & \multicolumn{1}{c}{Preserving $-$ swapped}
 & \multicolumn{1}{c}{Preserving $-$ no hint} & \multicolumn{1}{c}{Swapped $-$ no hint} \\
\midrule
\rowcolor{black!10}
\textbf{Qwen3} & 81.51 & \mci{+4.95}{1.43}{8.72} & \mci{-0.72}{-2.99}{1.63} & \mci{-5.66}{-10.03}{-1.43} \\
Qwen2 & 85.68 & \mci{+0.39}{-0.65}{1.56} & \mci{-1.37}{-3.06}{0.26} & \mci{-1.76}{-3.39}{-0.26} \\
LLaVA & 53.65 & \mci{+2.02}{0.13}{3.97} & \mci{+1.89}{-1.37}{5.27} & \mci{-0.13}{-4.17}{3.91} \\
GPT-4.1-mini & 79.69 & \mci{+0.91}{-1.69}{3.71} & \mci{+2.41}{0.07}{5.08} & \mci{+1.50}{-0.85}{3.84} \\
GPT-4o & 65.62 & \mci{+1.11}{-0.59}{2.86} & \mci{+2.34}{-0.78}{5.60} & \mci{+1.24}{-1.24}{3.78} \\
NVIDIA Llama & 24.22 & \mci{+3.06}{1.30}{4.88} & \mci{+19.53}{15.95}{23.05} & \mci{+16.47}{13.02}{19.92} \\
\bottomrule
\end{tabularx}
\caption{Hint benefit and harm relative to no hint on 48 Spatial pairs.
Shading highlights Qwen3's decomposition. Brackets give paired 95\% intervals.
NVIDIA's gains partly reflect fewer invalid-format answers.}
\label{tab:matched-effects}
\end{table*}

\subsection{Hint Rewriting and Task Reformulation}
\label{sec:task-design}

For a claim \textit{A left of B}, hints express the same relation as
\textit{A left of B} or \textit{B right of A}, and the opposite as
\textit{B left of A} or \textit{A right of B}. The latter two connect
the object swap to an equivalent relation reversal.
We compare field and sentence forms of each hint. A text-only probe
tests whether the hint agrees with the claim.

Verification asks \textit{Is the claim true in the image?} with Yes/No
options. Selection asks \textit{Where is the cup relative to the bowl in the image?}
with left/right options. Both keep the image, claim, and hint fixed
and use one A/B response. Task instructions and answer meanings change together.

Selection may also be easier without a hint. With accuracy $A_{t,h}$
for task $t$ and hint $h$, and $N$ denoting no hint, we adjust for this baseline:
\begin{align}
T_h ={}& (A_{\mathrm{select},h}-A_{\mathrm{verify},h}) \nonumber\\
       &-(A_{\mathrm{select},N}-A_{\mathrm{verify},N}).
\label{eq:task-hint}
\end{align}
Positive $T_h$ favors selection after adjustment.

\section{Experimental Setup}
\label{sec:setup}
\label{sec:study-data}

We evaluate ten models, detailed in Appendix~\ref{app:models}.
Material, Size, and Color each use 96 images, Spatial uses 48
opposite-truth pairs, and Holding and Attached-to use eight images each.

Attribute images come from GQA/Visual Genome annotations
\citep{hudson2019gqa,krishna2017visualgenome}. Each has two objects with
distinct values, giving two true claims and two false alternatives.
We use LLM-assisted review with manual checks by the authors
(Appendix~\ref{app:attribute-data}).

Spatial pairs use the same source and hold the left/right claim fixed
across opposite-truth images. The first 48 pairs passing result-blind
visibility, reference, and truth checks form both the label panel and
six-model hint study (Appendix~\ref{app:spatial-data}).

External-48 contains 48 previously used images disjoint from Spatial.
Qwen3 and mini answer two opposing claims per image, varying answer
mappings and question position
(Appendix~\ref{app:external-hints}).

RQ3 tests both models on a text-only probe and visual cohorts,
including a fixed follow-up of 24 images in 12 source pairs. Each image
has four claims and two answer mappings. Appendix~\ref{app:mitigation}
details the prompts, cohorts, and results. The follow-up retains all
nine pilot conditions on images and source pairs disjoint from the
earlier RQ3 cohorts.

A further comparison uses 17 Holding and 16 Attached-to images with plausible
participant types in both intended and swapped assignments. It compares
their interactions under real-image, gray-image, and text-only inputs
(Appendix~\ref{app:scope-details}).

Table~\ref{tab:fullmatrix} distinguishes local first-token (L1),
full-answer (LS), and API scores. A/B answers select claims in Attribute tasks and encode
Yes/No in Relation tasks, except the Spatial ablation uses Yes/No
directly. Baselines are protocol-specific.

Confidence intervals resample images or pairs with all prompts together
\citep{efron1994bootstrap}. In the original Spatial hint study, Qwen3's $D_C>0$ and $H>0$ are the
primary hint tests, with Holm correction. The External-48 tests are
specified in Appendix~\ref{app:external-hints}. RQ3 uses descriptive paired 95\%
bootstrap intervals over the 12 follow-up source pairs.
For all models, invalid-format answers count as errors
(Appendices~\ref{app:inference} and~\ref{app:matched}).

\section{RQ1: Effects of Role Labels}
\label{sec:panel}

To answer RQ1, we compare label responses across tasks, then identify
which components account for accuracy changes.

\subsection{Cross-Task Label Sensitivity}
\label{sec:cross-task}

Despite similar Color accuracy, LLaVA and Qwen3 have contrast estimates
of $+.057$ and $-.072$ (Table~\ref{tab:fullmatrix}).
Only LLaVA's interval excludes zero (Appendix~\ref{app:inference}).

\begin{figure}[t]
\centering
\includegraphics[width=\columnwidth]{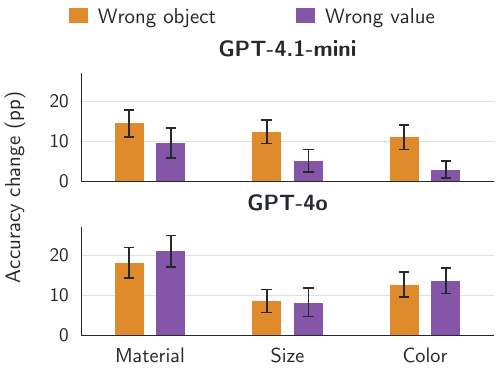}
\caption{Explicit labels affect the two error types differently.
Bars show accuracy changes from generic field names across six
model--task comparisons, with 95\% image-bootstrap intervals over
96 images per task. Appendix~\ref{app:hard-readout} gives condition accuracies.}
\label{fig:attribute-effects}
\end{figure}

Mini's NLL and generated-answer contrasts favor wrong-object
judgments across tasks (Figure~\ref{fig:attribute-effects}).
GPT-4o's readouts favor different error types on Material and Color,
with accuracy-contrast intervals spanning zero
(Appendix~\ref{app:hard-readout}).

On the eight-image Holding set, mini's generated-answer interaction is
$+7.81$ pp, although Both and Generic have identical accuracy.
Either component alone lowers accuracy. Here the positive interaction
reflects recovery from component losses, with no gain over Generic.

\subsection{Component Effects}
\label{sec:spatial}

The Spatial ablation separates these effects in accuracy
(Table~\ref{tab:cue-ablation}). Adding object-role labels to Relation
only helps Qwen3 by
4.69 pp but hurts Qwen2 by 3.39 pp. Both paired intervals exclude zero.
The cross-model difference is 8.07 pp $[4.17,12.11]$.

Gemma4 benefits mainly from the relation label, while LLaVA changes
little. Useful components thus depend on the model. Qwen3's best
field-based condition still falls below its ordinary-language reference.
Table~\ref{tab:cue-ablation} reports paired intervals for all six models.

\label{sec:association}

The compact Holding and Attached-to alternatives have implausible
participant types. A separate cohort keeps participant types plausible
in both assignments. Mean accuracy exceeds both blind controls, but the
intended-minus-swapped interaction intervals span zero for both relations.
Visual accuracy gains therefore need not imply a consistent role-label
interaction (Appendix~\ref{app:scope-details}).

The attribute results distinguish effects on object and value errors.
\textbf{For relations, object-role labels can help or hurt depending
on the accompanying relation label and the model.}

\section{RQ2: Benefits and Harms of Hints}
\label{sec:matched}

RQ1 examines labels within the claim being judged. Role information
can also be supplied as a separate hint. We therefore compare hinted
judgments with answering the original question alone, then examine
field order and paired errors.

\subsection{Help--Harm Decomposition}

Qwen3's preserving-over-swapped gap is 4.95 pp $[1.43,8.72]$.
Relative to no hint, preserving accuracy
changes by $-.72$ pp and swapped accuracy by $-5.66$ pp
(Table~\ref{tab:matched-effects}). The advantage therefore comes mainly from swapped-hint harm,
without a clear preserving-hint gain.

A role-preserving hint restates the question's roles.
It adds no visual evidence for deciding whether the claim is true.

Other models respond differently. Preserving effects range from gains
for GPT-4.1-mini to small losses for Qwen2. NVIDIA gives fewer invalid-format
answers with hints, so its larger gains combine better format compliance
with changes in answer choice (Table~\ref{tab:nvidia-controls}).

\begin{figure}[t]
\centering
\includegraphics[width=\columnwidth]{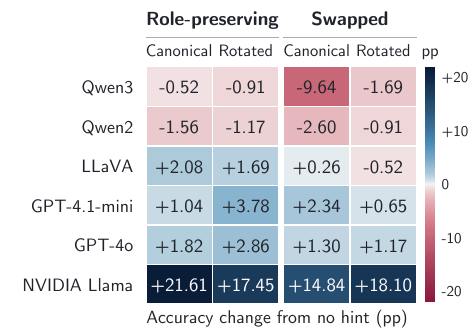}
\caption{Hint effects relative to no hint on 48 Spatial pairs (pp).
Blue indicates gains and red losses. Rotated averages two cyclic orders.
Paired intervals appear in Tables~\ref{tab:matched-effects}
and~\ref{tab:hint-order}.}
\label{fig:matched}
\end{figure}

\subsection{Content--Order Interactions}

Canonical order makes Qwen3's swapped hint more disruptive without a
comparable change in preserving-hint accuracy
(Figure~\ref{fig:matched}). The content-by-order interaction is
$8.33$ pp $[3.91,13.15]$.
Mini's interaction instead is $-4.43$ pp $[-7.94,-1.30]$.
Order effects thus depend on assignment and model. Token counts match
across hint variants in all five observable tokenizers.

Requiring both opposite-truth images to be answered correctly gives
the same qualitative pattern for Qwen3 (Table~\ref{tab:matched-cells}).

\begin{table}[t]
\centering
\small
\setlength{\tabcolsep}{3pt}
\newcommand{\ehci}[3]{$#1$\;{\scriptsize$[#2,#3]$}}
\begin{tabular*}{\linewidth}{@{\extracolsep{\fill}}lrr@{}}
\toprule
 & Qwen3 & GPT-4.1-mini \\
\midrule
No hint (\%) & 88.02 & 85.94 \\
\midrule
Preserving & \ehci{-1.39}{-3.99}{1.22} & \ehci{+0.87}{-1.22}{3.12} \\
Swapped & \ehci{-10.59}{-14.76}{-6.77} & \ehci{-2.26}{-4.86}{0.52} \\
Order & \ehci{+9.64}{3.91}{15.62} & \ehci{+7.03}{3.39}{11.20} \\
\bottomrule
\end{tabular*}
\caption{Role-hint effects on External-48 with the question first.
Preserving and Swapped give changes from no hint (pp),
averaged equally over three field orders. Order is the content-by-order
interaction (Eq.~\ref{eq:gamma-accuracy}). Brackets give paired 95\%
image-bootstrap intervals.}
\label{tab:external-hints}
\end{table}

\begin{figure*}[t]
\centering
\includegraphics[width=\textwidth]{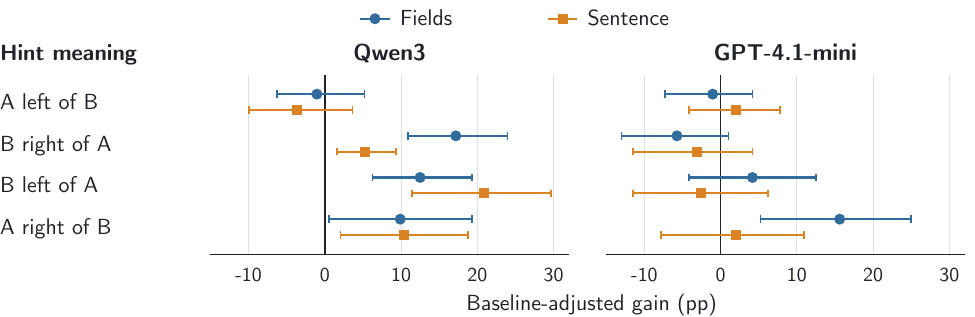}
\caption{Relation-selection gains after subtracting the no-hint task
difference (Eq.~\ref{eq:task-hint}). All eight hint conditions are shown.
Points show estimates and lines show paired 95\% bootstrap intervals over 12 source pairs
(24 images, 192 judgments per condition). Hint meanings refer to A left of B.
Table~\ref{tab:mitigation-cells} gives complete accuracies and error counts.}
\label{fig:mitigation}
\end{figure*}

\subsection{Replication and Error Patterns}
\label{sec:external-hints}

On External-48, Qwen3 again shows swapped-hint harm without
a clear preserving gain (Table~\ref{tab:external-hints}). Its order
interaction remains positive, whereas mini's changes sign across
cohorts. Qwen3's pattern also holds under both field-order weighting schemes
(Appendix~\ref{app:order-weight}).

For Qwen3, canonical swapped hints increase the number of opposing
claim pairs answered No to both from 19 to 49 out of 96.
Both-answer reversals remain at zero. The extra errors mainly reject
true claims rather than reverse both judgments
(Table~\ref{tab:external-patterns}). Disagreement between the hint and
question may encourage rejection of the claim.
Question-last presentation lowers mini's canonical swapped-hint accuracy.
Qwen3's interval spans zero.
Repeating the question yields no clear reduction in harm
(Appendix~\ref{app:external-hints}).

\begin{samepage}
Together, these comparisons show that \textbf{Qwen3's role-preserving
advantage mainly reflects swapped-hint harm, amplified by canonical
field order.}

\end{samepage}

\section{RQ3: Mitigating Hint-Induced Errors}
\label{sec:mitigation}

RQ2's extra rejections motivate changes to hints and the judgment task.
We test clearer wording to aid role interpretation and direct selection
of the depicted relation.

\subsection{Meaning-Preserving Hint Rewriting}

In the text-only probe, both models answer all 32 judgments correctly with sentence hints,
improving over canonical fields.
Yet better text matching does not consistently improve visual judgments.
In the 24-image follow-up, sentence rewriting raises Qwen3's verification
accuracy by 14.58 pp for \textit{B right of A}, but lowers it by 21.35 pp
for \textit{B left of A}. The initial visual cohort shows the same
opposing pattern. Mini's changes vary across hints and cohorts
(Appendix~\ref{app:mitigation}). Clearer hints alone do not reliably
resolve visual errors.

\subsection{Verification versus Selection}

Figure~\ref{fig:mitigation} shows all eight hinted contrasts after
adjusting for the no-hint task difference.
Positive values indicate a larger selection advantage with the hint
than without it.
The clearest shared recovery occurs for \textit{A right of B} in field
form. The contrasts $T_h$ are 9.90 pp $[.52,19.27]$ for Qwen3 and
15.62 pp $[5.21,25.00]$ for
mini. These gains remain positive for both models
after omitting any one source pair
(Appendix~\ref{app:existing-sensitivity}).

In this condition, Qwen3 corrects 26 judgments without introducing
new errors. Mini corrects 55 judgments and introduces 22 errors.
Both remain below their unhinted selection accuracy. More opposing
claims are answered correctly together, but answers become more
sensitive to the answer mapping. Other hints also produce gains and
losses for mini (Appendix~\ref{app:mitigation}).
\textbf{Relation selection can partially recover judgments
disrupted by role hints, whereas better text matching does not ensure
more accurate visual judgments.}

\section{Discussion}

Component ablations identify useful labels within a structured prompt.
No-hint comparisons establish whether adding role information helps.
When user-supplied text must remain in the input, RQ3's task change
recovers some disrupted judgments. Omitting the hint still yields
higher accuracy in the main recovery condition.
These findings favor choosing labels for a specific model and task
over adding structure by default. The observed error pattern can then guide whether to revise the hint's wording or change the judgment task.

These comparisons separate understanding a hint from using it in a
visual judgment. Text-only matching tests interpretation of the stated
roles. Visual evaluation then measures which errors an intervention
corrects and which it introduces.
In practice, this suggests testing role interpretation before changing
the visual task. Poor text matching motivates clearer wording.
If interpretation improves but visual errors persist, task reformulation
offers a different route to test on the same inputs.

\paragraph{Application to answer checking.}
\label{sec:reliability}
Each strategy ranks the original answer by its disagreement with three
additional checks, without revising it.
With wording, answer labels, and query counts matched, role-explicit
checks improve error ranking for GPT-4.1-mini and GPT-4o, with no clear
gain for the other four models (Appendix~\ref{app:reliability}).
This distinguishes two uses of role information: correcting judgments
and detecting unreliable answers. Correction is assessed by changes in accuracy. Checking is assessed by how well it ranks errors in the original answers. In practice, the ranking could help prioritize answers for human review when reviewing every prediction is impractical.

\section{Conclusion}

We examined how role labels and auxiliary hints shape visual judgments.
Matched comparisons across tasks show that labels can affect object
and value errors differently, even when ordinary accuracy is similar.
For relations, useful label components depend on the model and on which
other labels are present. No-hint comparisons show that swapped-hint
harm can produce a role-preserving advantage without a preserving-hint
gain.

These diagnoses also guide intervention. Clearer hint wording improves
text matching but does not consistently improve visual judgments.
Relation selection partially recovers disrupted judgments, with benefits
that vary across hints and models. Together, the studies connect prompt
components, error patterns, and intervention outcomes. They guide the choice
of role cues and judgment tasks through observed error patterns, rather
than adding structure by default.

\section*{Limitations}

Our curated English-language cohorts limit generalization. Small
association sets yield wide intervals (Appendix~\ref{app:inference}),
and model-specific readouts limit comparisons of score magnitudes.

Both RQ2 image sets were used previously. External-48 also changes
question pairing. The answer-checking studies and
Table~\ref{tab:cue-ablation}'s precision analysis follow earlier results.
RQ3 covers two models, with 12 source pairs in its fixed follow-up.

Without a length-matched neutral hint, changes from no hint do not
isolate role semantics from the influence of added text. RQ3 changes
task instructions and answer meanings together, leaving their separate
contributions unresolved.

Our findings characterize behavioral responses to prompt interventions, rather than internal model mechanisms.

\section*{Ethical Considerations}

Any release will respect source licenses and privacy, using source
identifiers, links, attribution, and permitted annotations rather than
redistributing original images or private annotated renderings.

\bibliography{references_arxiv}
\appendix
\raggedbottom

\section*{Appendix Contents}
\begingroup
\small
\setlength{\tabcolsep}{3pt}
\renewcommand{\arraystretch}{1.12}
\newcommand{\appentry}[2]{\ref{#1} & \hyperref[#1]{#2} & \pageref{#1}\\}
\begin{tabularx}{\columnwidth}{@{}lYr@{}}
 & & Page \\
\appentry{app:related}{Extended Related Work}
\appentry{app:models}{Models and Inference}
\appentry{app:attribute-data}{Attribute Data and Matched Prompts}
\appentry{app:spatial-data}{Spatial Cohort and Auxiliary Controls}
\appentry{app:matrix}{Shared-Panel Scoring}
\appentry{app:scope-details}{Additional Cross-Task Evidence}
\appentry{app:inference}{Statistical Analysis}
\appentry{app:prediction-budget}{Ordinary-Query and Nonlinear Predictive Controls}
\appentry{app:matched}{Auxiliary-Hint Evaluation}
\appentry{app:external-hints}{Role Hints on External-48}
\appentry{app:mitigation}{Hint Rewriting and Response Tasks}
\appentry{app:reliability}{Answer Checking: Matched and Order-Only Comparisons}
\appentry{app:factual}{Attribute Judgments with Factual Histories}
\end{tabularx}
\par
\endgroup

\section{Extended Related Work}
\label{app:related}

\paragraph{Visual Binding and Compositional Evaluation.}
Compositional evaluation asks whether models preserve the structure of
a scene, not only whether they recognize familiar objects and words.
Diagnostic tasks test attributes and relations through controlled scenes
or minimally different descriptions
\citep{johnson2017clevr,thrush2022winoground,yuksekgonul2023aro,parcalabescu2022valse,zhao2022vlchecklist}.
VSR covers diverse spatial relations, whereas CREPE varies
compositional novelty and complexity \citep{liu2023vsr,ma2023crepe}.
For these tests to isolate visual understanding, counterexample
construction must also account for linguistic shortcuts
\citep{hsieh2023sugarcrepe,ma2024sade}.
The same need to preserve assignments extends to objects that share
features, persist through time, or are referenced across modalities
\citep{campbell2025binding,saravanan2024velociti,alonso2025mate}.
Representation-level studies complement these behavioral tests by
examining how models bind entities in context and how external cues
affect those bindings \citep{feng2024binding,hasani2025groundingids}.
Relation-guided attention provides a further way to intervene on the
use of these assignments \citep{bukkapatnam2026reground}.
SpatialRGPT uses region-level training and optional depth inputs
to improve spatial reasoning \citep{cheng2024spatialrgpt}.
Our comparisons examine the behavioral effects of making these roles
explicit in the prompt.

\paragraph{Prompt Sensitivity and Textual Interference.}
Even when meaning is preserved, prompt formatting can substantially
change LLM performance, with different effects across models
\citep{sclar2024formatspread}. In visual tasks,
additional text can override visual evidence, with effects that vary
by task, model, and input order
\citep{deng2025words,pezeshkpour2025mixed}.
VLind-Bench controls for basic perception failures when testing
language priors \citep{lee2025vlind}.
In answer evaluation, judges can also favor informative responses
despite visual contradictions \citep{zou2026informativeness}.
Related work also shows why the truth of added text is not enough
to establish its usefulness. Conversation history
can carry earlier errors forward, and even true hints can mislead
\citep{zhong2024snowball,shafiei2026truthtrap}.
Expressions of confidence or uncertainty also affect responses, making
presentation another relevant factor \citep{zhou2023grey}.
These findings motivate separating what a hint says from how it is
presented. We vary the hint's agreement with the question's roles and
its field order independently. The no-hint baseline then separates
help from harm.

\paragraph{Visual Judgment and Correction.}
POPE tests object presence through binary questions \citep{li2023pope}.
VQAScore uses the probability of a yes answer to measure image--text
alignment \citep{lin2024vqascore}.
These tasks differ from comparing two candidate relations.
For correction, visual contrastive decoding contrasts predictions
from original and distorted images \citep{leng2024vcd}.
TACO instead checks atomic claims through paraphrases and confidence
aggregation before refining answers \citep{liu2026taco}.
We examine role wording and judgment format without changing model
weights or decoding.

\paragraph{Uncertainty Estimation and Evidence Use.}
Uncertainty estimation uses self-assessment and variation across
responses as evidence about answer reliability
\citep{xiong2026enhancing, xiong2025deliberate, bao2026driftbench, zhang2026semantic, zhang2026oracle, shi2026confidence, shi2026sage}.
VLMs' verbalized confidence is also often miscalibrated across tasks
\citep{xuan2025vcap}.
However, agreement alone need not rule out a shared mistake.
Related work distinguishes ambiguous or insufficient inputs from failures
to use available evidence
\citep{cole2023ambiguous,joren2025sufficient,wagner2026twoaxes,eisenschlos2024selectively}.
For visual tasks, acquiring an informative view changes the evidence
available to the model \citep{zhang2026seeing}. In contrast, our matched
study holds the image fixed and varies how object and relation roles are
stated.

\paragraph{Selective Visual Question Answering.}
For selective VQA, useful uncertainty signals should improve the
trade-off between retaining answers and making errors
\citep{whitehead2022reliable,khan2024consistency}. One way to obtain these
signals is to vary object references and question wording to target
binding errors \citep{yang2025sources}. This motivates examining which
role information an answer check should make explicit. Presentation order
is a separate influence on responses \citep{ok2026order}, so our hint
study varies role content and order independently. Our answer-checking
comparison matches query counts, wording changes, and the disagreement
score to test whether role-explicit hints improve error detection.
\section{Models and Inference}
\label{app:models}

The local models are
\href{https://huggingface.co/llava-hf/llava-v1.6-mistral-7b-hf}{LLaVA-v1.6-Mistral-7B},
\href{https://huggingface.co/Qwen/Qwen2-VL-7B-Instruct}{Qwen2-VL-7B-Instruct},
\href{https://huggingface.co/Qwen/Qwen3-VL-8B-Instruct}{Qwen3-VL-8B-Instruct},
\href{https://huggingface.co/google/gemma-4-E4B-it}{Gemma-4-E4B-it},
\href{https://huggingface.co/OpenGVLab/InternVL3_5-8B-Instruct}{InternVL3.5-8B-Instruct},
\href{https://huggingface.co/zai-org/GLM-4.1V-9B-Thinking}{GLM-4.1V-9B-Thinking},
and \href{https://huggingface.co/microsoft/Phi-4-multimodal-instruct}{Phi-4-multimodal-instruct}.
The API models are GPT-4.1-mini (2025-04-14), GPT-4o (2024-11-20),
and NVIDIA-hosted Llama-3.2-11B-Vision-Instruct.

Table~\ref{tab:fullmatrix} identifies first-token, full-answer, and API
score readouts. Appendix~\ref{app:matrix} defines their aggregation.
Generated-answer studies use greedy decoding and A/B response parsing.
Each protocol specifies its output limits.

For GLM-4.1V-Thinking, candidate answers are teacher-forced directly
after the prompt, without generating an intermediate reasoning trace.
The reported scores thus evaluate direct answers, not extended reasoning.
Gemma4's Attribute rows use first-token A/B scores with thinking disabled.
Their ordinary accuracies are 38.0\%, 42.7\%, and 46.1\% for Material,
Size, and Color. These runs do not explain the low accuracies.

In the RQ3 follow-up, Qwen3 runs in bfloat16 on one H100-80GB GPU,
using PyTorch 2.8.0, CUDA 12.8, and Transformers 4.57.6.
Images use the checkpoint's standard processor.
GPT-4.1-mini uses the API's high-detail image setting.
Both decode greedily, capped at eight tokens.

The linked model cards list Apache 2.0 for LLaVA, Qwen2, Qwen3,
Gemma4, and InternVL3.5, and MIT for GLM-4.1V and Phi-4.
Llama uses the
\href{https://huggingface.co/meta-llama/Llama-3.2-11B-Vision-Instruct}{Llama 3.2 Community License}.
Hosted API use is additionally subject to the providers' service terms.
\newsavebox{\promptbody}
\newenvironment{promptbox}[1]{%
  \par\addvspace{6pt}\noindent
  \begin{minipage}{\linewidth}
  \small\rmfamily\raggedright
  \textbf{#1}\par\nobreak\vspace{3pt}
  \setlength{\fboxsep}{4pt}%
  \setlength{\fboxrule}{0.35pt}%
  \begin{lrbox}{\promptbody}%
  \begin{minipage}{\dimexpr\linewidth-2\fboxsep-2\fboxrule\relax}
  \small\rmfamily\raggedright
  \setlength{\parindent}{0pt}%
  \setlength{\parskip}{3pt}%
  \ignorespaces
}{%
  \end{minipage}%
  \end{lrbox}%
  \noindent\fcolorbox{black!45}{white}{\usebox{\promptbody}}\par
  \end{minipage}\par\addvspace{6pt}%
  \ignorespacesafterend
}
\section{Attribute Data and Matched Prompts}
\label{app:attribute-data}

We use GQA annotations derived from Visual Genome. Visual Genome lists
\href{https://visualgenome.org/about}{CC BY 4.0} for its dataset.
Source photographs retain their respective rights, separate from the
annotation license.

\paragraph{Construction.}
Material, Size, and Color each contain 96 natural images selected from
GQA/Visual Genome training annotations. Each image supplies two
distinguishable objects, $o_0,o_1$, with distinct attribute values,
$v_0,v_1$. The two annotated assignments are true, and exchanging
their values must produce two false claims. This forms the quartet
\begin{equation}
\begin{array}{c|cc}
 & v_0 & v_1\\\hline
o_0 & \mathrm{true} & \mathrm{false}\\
o_1 & \mathrm{false} & \mathrm{true}
\end{array}
\end{equation}
For the anchor $(o_0,v_0)$, $(o_1,v_0)$ is the wrong-object alternative
and $(o_0,v_1)$ the wrong-value alternative. The anchor $(o_1,v_1)$
provides the corresponding reverse comparisons. Thus both error types
use the same objects, values, and true claims.

LLM-assisted review and author checks assess object visibility, unique
reference, and the truth of all four claims. Disagreements are resolved
without access to evaluated model responses. The first 96 qualifying
images in a fixed order form each task. Material and Size use 288
initial candidates in total. Color uses two batches of 144 candidates,
both reviewed before model evaluation.

\paragraph{Prompt templates.}
Each comparison tests two stems in separate queries:
\emph{Which structured record is supported by the image?} and
\emph{Choose the structured record that correctly describes the image.}
One stem is inserted into the template below.
\begin{promptbox}{Attribute judgment}
\textit{[Stem]}\\
A. \textit{[First record]}\\
B. \textit{[Second record]}\\
Your entire response must be exactly one character: A~or~B.
\end{promptbox}
Bracketed items are placeholders. The first and second records are the
true claim and one false alternative, tested in both option orders.
Explicit records use
\texttt{(object=\textit{[object]}; attribute=\textit{[value]})}.
Generic records use
\texttt{(part1=\textit{[object]}; part2=\textit{[value]})}.
The field values and their order are identical. Each label condition
therefore contains two true anchors, two error types, two stems, and
two option orders. Appendix~\ref{app:matrix} explains aggregation for
probability scores and generated answers.
\section{Spatial Cohort and Auxiliary Controls}
\label{app:spatial-data}

\subsection{Data construction and reuse}

The Spatial cohort uses GQA/Visual Genome scene-graph
candidates \citep{hudson2019gqa,krishna2017visualgenome}, excluding the full
144-image development set. We select horizontal left/right candidates
whose object-box centers are separated by at least 18\% of image width.
We pair the same literal proposition
across images with opposite truth values. This produces 120 candidate
pairs (240 images) for result-blind visual review.

LLM-assisted review checks object visibility, unique reference, box
alignment, and relation truth, with a further review of disputed items.
Of 211 qualifying images, 91 complete pairs remain. The first 48 complete
pairs in the fixed order form the 96-image cohort, balanced between
48 true and 48 false propositions. No model response enters admission.
This cohort is disjoint from the earlier training and test
images. Subsequent auxiliary-hint and checking studies reuse its 48 pairs.

The separate type-valid association cohort uses LLM-assisted review
and author checks of 64 candidates. A fixed agreement rule selects
17 Holding and 16 Attached-to images. The eight-image association
cohorts are constructed separately.

\subsection{Coherent versus malformed records}
\label{app:spatial-controls}

The malformed-record control retains the same values $S,R,O$ but cycles
their field names. Besides \texttt{subject=S; relation=R; object=O}, it uses
\texttt{object=S; subject=R; relation=O} and
\texttt{relation=S; object=R; subject=O}. These malformed records place
a predicate in an entity field and an entity in a relation field.

\begin{figure}[t]
\centering
\includegraphics[width=\linewidth]{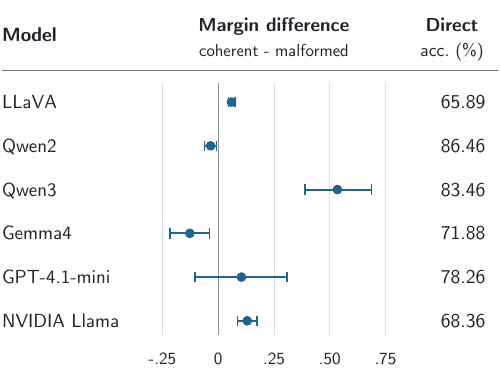}
\caption{Coherent-minus-malformed answer margins across six models,
with 95\% pair-bootstrap intervals. Direct-judgment accuracy is shown
alongside. All results use the same 96 images in 48 pairs.}
\label{fig:spatial}
\end{figure}

The contrast measures sensitivity to coherent field structure.
Figure~\ref{fig:spatial} compares this response with direct-judgment
accuracy across all six models.
Qwen2 answers separately worded questions
more accurately than Qwen3 (.865 versus .835). The coherent-minus-malformed
mean answer margin is $-.034$ for Qwen2 and $+.535$ for Qwen3, with
95\% intervals $[-.061,-.008]$ and $[.390,.688]$
(Figure~\ref{fig:spatial}). The contrast includes both role assignment
and record interpretability.

Across the Spatial panel, the mean coherent-minus-malformed advantage is
$+.111$ with a 95\% interval of $[.063,.160]$. Models also use visual
evidence: mean accuracy is 78.56\% with real images and 50\% under each
of two blind prompt variants that replace the image with gray. Each pair
asks the same question but has opposite correct answers, so an identical
image-blind response gets one of the two right.

\paragraph{Illustration credit.}
The photograph in Figure~\ref{fig:role-prompts} is by
\href{https://unsplash.com/photos/two-dogs-running-in-the-grass-near-a-tree-P7ngl_WZyKI}{Caleb Bennetts on Unsplash},
used under the \href{https://unsplash.com/license}{Unsplash License}.
It was cropped and annotated by the authors.

\section{Shared-Panel Scoring}
\label{app:matrix}

Table~\ref{tab:fullmatrix} reports all 60 model--task cells.
L1 denotes local first-step scoring, LS local full-sequence scoring, and API
an API-based readout. Each task pairs ordinary accuracy with a response
contrast computed from answer scores. All 60 cells use the
task-specific definitions in Eqs.~\ref{eq:gamma} and~\ref{eq:gamma-attribute}.
Generated-answer comparisons are reported separately in
Appendix~\ref{app:hard-readout}.
$-.000$ denotes a rounded small negative value.

\paragraph{Ordinary judgments and aggregation.}
Ordinary accuracy scores the original questions rather than the Generic
field condition. For score-based Attributes, each comparison pools
truth-oriented A/B log-odds over two templates and both answer positions.
If the mean margin is $\bar z$, correctness is one when $\bar z>0$
and zero otherwise. The binary candidate-normalized NLL is
$\log(1+\exp(-\bar z))$.
Accuracy averages the four comparison indicators, covering two anchors
and two error types, within each image. Each of the 96 images receives
equal weight. The NLL contrasts likewise pool margins before computing
loss, separately for each label condition. Explicit-minus-generic
loss changes are averaged over anchors within each error type.

For score-based Relations, correctness instead thresholds each
ordinary-criterion margin before averaging within clusters.
Generated-answer accuracy scores each returned A/B string before
averaging. Images, or Spatial image pairs, receive equal weight in
both cases.

The six supplementary score collections use the same first-token readout
with one- or eight-token generation limits. Images, prompts, answer mappings,
and model snapshots are unchanged. Every primary comparison has both A/B
scores.

\subsection{Label effects in generated answers}
\label{app:hard-readout}

Answer scores can change without changing the chosen answer. We therefore
also measure how field labels affect correctness by scoring generated A/B
answers directly. The comparisons retain the two task designs: differences
between error types for Attributes, and interactions between labels for
Relations. Tables~\ref{tab:hard-readout} and~\ref{tab:hard-conditions}
report the summary contrasts and the conditions that produce them.

Ordinary accuracy averages individual answer correctness within each image
or image pair, then weights these independent units equally. Attribute
accuracy therefore differs from the score-based readout, which thresholds
averaged answer-score margins. For an Attribute error type $e$, let
$d_e=a_{\mathrm{explicit},e}-a_{\mathrm{generic},e}$. The accuracy
interactions are
\begin{align*}
\gamma^{\rm acc}_{\rm attr}&=d_{\rm obj}-d_{\rm val},\\
\gamma^{\rm acc}_{\rm rel}&=a_{11}-a_{10}-a_{01}+a_{00}.
\end{align*}
Here, $a$ denotes accuracy in the specified label condition, and
obj/val denote wrong-object and wrong-value alternatives. The Attribute
contrast is positive when explicit labels help wrong-object judgments
more than wrong-value judgments. Its error-cost counterpart has the
opposite sign. The Relation contrast is positive when the benefit of an
object-role component is greater with the relation label than without it.
We report these accuracy contrasts in percentage points.

\begin{table*}[t]
\centering
\small
\newcommand{\hci}[3]{\shortstack{$#1$\\{\scriptsize$[#2,#3]$}}}
\begin{tabular*}{\linewidth}{@{\extracolsep{\fill}}llrcc@{}}
\toprule
Model & Task & $N$ & Ordinary accuracy (\%) & $\gamma^{\rm acc}$ (pp) \\
\midrule
GPT-4.1-mini & Spatial & 48 & \hci{77.34}{73.31}{81.25} & \hci{-0.52}{-3.39}{2.34} \\[5pt]
 & Holding & 8 & \hci{81.25}{65.62}{95.31} & \hci{+7.81}{1.56}{14.06} \\[5pt]
 & Attached-to & 8 & \hci{87.50}{71.88}{98.44} & \hci{+6.25}{-7.81}{25.00} \\[5pt]
 & Material & 96 & \hci{94.86}{92.19}{97.20} & \hci{+4.95}{0.91}{8.85} \\[5pt]
 & Size & 96 & \hci{81.71}{76.82}{86.33} & \hci{+7.16}{3.12}{11.07} \\[5pt]
 & Color & 96 & \hci{96.42}{93.95}{98.50} & \hci{+8.07}{4.82}{11.20} \\
\midrule
GPT-4o & Material & 96 & \hci{94.40}{91.60}{96.88} & \hci{-2.86}{-7.29}{1.43} \\[5pt]
 & Size & 96 & \hci{79.62}{75.39}{83.79} & \hci{+0.26}{-3.91}{4.56} \\[5pt]
 & Color & 96 & \hci{95.70}{92.90}{98.05} & \hci{-0.91}{-4.69}{2.86} \\[5pt]
 & Spatial & 48 & \hci{67.71}{64.06}{71.22} & \hci{-2.73}{-7.16}{1.56} \\
\bottomrule
\end{tabular*}
\caption{Label effects measured from generated answers. Ordinary accuracy
summarizes the original questions. $\gamma^{\rm acc}$ measures how label
gains differ between error types (Attributes) or depend on the other label
(Relations). Brackets are
descriptive 95\% cluster-bootstrap intervals. $N$ counts image pairs for
Spatial and images otherwise. All rows reuse the original cohorts.
Table~\ref{tab:hard-conditions} gives the underlying condition accuracies.}
\label{tab:hard-readout}
\end{table*}

\begin{table*}[t]
\centering
\small
\newcommand{\hcci}[3]{\shortstack{$#1$\\{\scriptsize$[#2,#3]$}}}
\begin{tabular*}{\linewidth}{@{\extracolsep{\fill}}llrrrrcc@{}}
\toprule
\multicolumn{8}{l}{\textbf{(a) Attributes: generic (G) versus explicit (E) field names}} \\
\addlinespace[3pt]
 & & \multicolumn{2}{c}{Wrong object} & \multicolumn{2}{c}{Wrong value} & & \\
\cmidrule(lr){3-4}\cmidrule(lr){5-6}
Model & Task & G & E & G & E & $d_{\rm obj}$ & $d_{\rm val}$ \\
\midrule
GPT-4.1-mini & Material & 71.22 & 85.68 & 82.81 & 92.32 & \hcci{+14.45}{11.07}{17.84} & \hcci{+9.51}{5.86}{13.28} \\[5pt]
 & Size & 61.46 & 73.70 & 73.44 & 78.52 & \hcci{+12.24}{9.38}{15.23} & \hcci{+5.08}{2.34}{7.94} \\[5pt]
 & Color & 78.65 & 89.58 & 92.84 & 95.70 & \hcci{+10.94}{7.94}{14.06} & \hcci{+2.86}{0.78}{5.08} \\[5pt]
GPT-4o & Material & 64.19 & 82.29 & 68.75 & 89.71 & \hcci{+18.10}{14.32}{21.88} & \hcci{+20.96}{17.06}{24.87} \\[5pt]
 & Size & 60.16 & 68.62 & 64.32 & 72.53 & \hcci{+8.46}{5.73}{11.46} & \hcci{+8.20}{4.69}{11.85} \\[5pt]
 & Color & 75.13 & 87.76 & 79.69 & 93.23 & \hcci{+12.63}{9.51}{15.89} & \hcci{+13.54}{10.42}{16.80} \\
\midrule
\multicolumn{8}{l}{\textbf{(b) Relations: object-role and relation labels, separately and together}} \\
\addlinespace[3pt]
Model & Task & Generic & Object-role & Relation & Both & \shortstack{Object-role\\effect} & \shortstack{Relation\\effect} \\
\midrule
GPT-4.1-mini & Spatial & 65.23 & 66.67 & 64.58 & 65.49 & \hcci{+1.17}{-0.91}{3.32} & \hcci{-0.91}{-2.80}{0.98} \\[5pt]
 & Holding & 85.94 & 81.25 & 82.81 & 85.94 & \hcci{-0.78}{-3.91}{3.91} & \hcci{+0.78}{-3.91}{3.91} \\[5pt]
 & Attached-to & 84.38 & 82.81 & 82.81 & 87.50 & \hcci{+1.56}{-3.12}{8.59} & \hcci{+1.56}{0.00}{3.91} \\[5pt]
GPT-4o & Spatial & 53.12 & 54.04 & 56.90 & 55.08 & \hcci{-0.46}{-3.12}{2.34} & \hcci{+2.41}{0.13}{4.69} \\
\bottomrule
\end{tabular*}
\caption{Condition accuracies (\%) and component effects (pp) underlying
Table~\ref{tab:hard-readout}. Attribute effects are explicit minus generic.
Relation effects average the change from adding a label over the presence
and absence of the other label. Effect intervals are descriptive 95\%
cluster-bootstrap intervals. Differences are computed before rounding.}
\label{tab:hard-conditions}
\end{table*}

Mini's Color accuracy contrast ($+8.07$ pp, 95\% CI $[4.82,11.20]$)
and NLL contrast ($-.745$) both favor wrong-object judgments:
gains and costs have opposite signs. For GPT-4o, the NLL and accuracy
readouts instead favor different error types on Material and Color.
The accuracy contrasts are $-2.86$ pp $[-7.29,1.43]$ and
$-0.91$ pp $[-4.69,2.86]$, respectively. Both CIs span zero,
as do mini's Attached-to and GPT-4o's Size CIs.

\paragraph{Collection and estimation.}
Generated answers provide a complementary readout for mini's three
Relation tasks and GPT-4o's three Attribute tasks. Mini Color and GPT-4o
Spatial additionally compare readouts against the earlier score-based
estimates. These two cells were selected to cover both task designs before
viewing the new responses. All eight reuse the admitted images, original
question strings, and answer mappings.

We use snapshots GPT-4.1-mini (2025-04-14) and
GPT-4o (2024-11-20), with temperature zero and an eight-token output limit.
The parser accepts
only A/B with optional surrounding whitespace and a final period or
closing parenthesis. All 26,752 requests have responses across two
collection batches. Four returned
strings are invalid, three for mini Color and one for GPT-4o Size,
and count as incorrect. Intervals use 20,000 cluster-bootstrap draws,
keeping every condition within its image or pair together.

To complete the two-model Attribute comparison, we also score mini's
Material and Size answers with the same parser, image weighting,
and bootstrap procedure. Each task has 4,608 core answers over 96 images.
These earlier runs used the same model snapshot and prompts but a
one-token output limit, rather than eight. One Material answer and nine
Size answers are invalid and count as incorrect. The within-task label
comparisons are matched. Cross-model comparisons also differ in
execution batch and output limit.

\subsection{Average component effects}

The average relation-label effects are
\begin{align*}
\alpha&=\tfrac12[(z_{10}-z_{00})+(z_{11}-z_{01})],\\
\beta&=\tfrac12[(z_{01}-z_{00})+(z_{11}-z_{10})].
\end{align*}
Thus $z_{11}-z_{00}=\alpha+\beta$, which does not determine
$\gamma_{\rm rel}$. For attributes, the average label-induced NLL
change is $\alpha_{\rm attr}=(B+F)/2$.
On compact Holding, Qwen2 has higher ordinary accuracy than Qwen3
(.844 versus .781), but a smaller score interaction
($+.104$ versus $+.684$). Thus the ranking by interaction differs
from the ranking by ordinary accuracy on these eight images.
In the 96-image Spatial ablation, Qwen3's Brier score falls from .281
with generic fields to .209 with both components.
The 24-image Spatial development study also places Both first for
Qwen3, with a 17.19 pp gain over generic fields.
\section{Additional Cross-Task Evidence}
\label{app:scope-details}

\subsection{Cross-Template Prediction}

The compact Holding and Attached-to studies keep a literal proposition fixed
while changing role fields and equivalent templates. In the six-model
confirmation, adding the first template's interaction reduced
leave-one-image-out prediction MSE for the second template by 24.07\% and
24.63\%, respectively. The prediction target is the interaction under
the second template on the same scenes.

The alternative role fields in these eight-image studies are malformed
or type-incongruent. Their contrast with coherent fields combines role
assignment with structural plausibility. We therefore test a separate
cohort with plausible participant types under both assignments.

\subsection{Type-Valid Role Assignments}

A separate study uses 17 Holding and 16 Attached-to
images, type-valid intended and swapped assignments, independently worded
direct criteria, and real-image, gray-carrier, and proposition-only
conditions. Five of six specified models have complete usable responses.
NVIDIA's collection stopped after eight probes because a response did not
meet the A/B format, leaving no complete model-level estimate.

\begin{table}[t]
\centering
\small
\setlength{\tabcolsep}{4pt}
\begin{tabular*}{\linewidth}{@{\extracolsep{\fill}}lrr@{}}
\toprule
 & Holding & Attached-to \\
\midrule
Images & 17 & 16 \\
Real-image accuracy & 71.76\% & 81.25\% \\
Gray-carrier accuracy & 40.00\% & 63.75\% \\
Proposition-only accuracy & 51.18\% & 58.13\% \\
\midrule
Role contrast & $+.090$ & $-.616$ \\
95\% image CI & $[-.630,.799]$ & $[-1.478,.319]$ \\
\bottomrule
\end{tabular*}
\caption{Separate type-valid association study. Role contrast is
intended-minus-swapped $\gint$, with an image-bootstrap interval, averaged
equally over five complete fixed models (one API, four local). Accuracy
comparisons are descriptive. Model-level results are in
Table~\ref{tab:assocmodels}.}
\label{tab:association}
\end{table}

Real-image performance exceeds both blind conditions in each family
(Table~\ref{tab:association}). The intended-minus-swapped interaction
is near zero for Holding and negative for Attached-to, with both
intervals spanning zero. The joint criterion requires positive role
and incremental-prediction effects in both families and is not met.
Prediction results likewise vary by family.
Defining improvement as baseline minus augmented MSE, Holding yields
$-.006672$ (95\% CI $[-.016264,-.000854]$), whereas Attached-to yields
$+.000113$ ($[-.008472,.007408]$). Here MSE measures squared probability
error, whereas Spatial uses squared margins.
Secondary API-only and local-only estimates are negative in both
families (Appendix~\ref{app:inference}).

Individual-model descriptives illustrate task dependence. Qwen3's role
contrast is $+4.107$ for Holding but $-1.750$ for Attached-to.
The two families use different images.
Appendix~\ref{app:association} gives all model-level estimates.

The two studies answer complementary questions. Cross-template prediction
shows that label responses recur across wordings. Type-valid comparisons
test preference for a particular role assignment and reveal substantial
variation across models and relations. Isolating the effect of control
quality would require comparing both control designs on the same images.

\subsection{Model-level association results}
\label{app:association}

Table~\ref{tab:assocmodels} shows the model differences behind
Table~\ref{tab:association}. The role contrast is intended-minus-swapped
interaction. API and local readouts are identified for each model.
Each of the five complete models contributes 1,782 probes across the
33 images and all specified conditions.

\begin{table*}[t]
\centering
\small
\begin{tabular*}{\linewidth}{@{\extracolsep{\fill}}llrrrr@{}}
\toprule
 & & \multicolumn{2}{c}{Holding} & \multicolumn{2}{c}{Attached-to} \\
\cmidrule(lr){3-4}\cmidrule(l){5-6}
Model & Readout & Accuracy & Role contrast & Accuracy & Role contrast \\
\midrule
GPT-4.1-mini & API & 79.41\% & $-1.967$ & 93.75\% & $-.719$ \\
Gemma4 & Local & 70.59\% & $-1.371$ & 78.13\% & $-.148$ \\
LLaVA & Local & 70.59\% & $-.026$ & 75.00\% & $-.090$ \\
Qwen2 & Local & 52.94\% & $-.290$ & 78.13\% & $-.375$ \\
Qwen3 & Local & 85.29\% & $+4.107$ & 81.25\% & $-1.750$ \\
\bottomrule
\end{tabular*}
\caption{Real-image results for the five complete models in the separate
type-valid cohort. Primary role estimates equally weight these models.}
\label{tab:assocmodels}
\end{table*}
\section{Statistical Analysis}
\label{app:inference}

\subsection{Shared-panel inference}

The shared-panel analysis uses 10,000 percentile-bootstrap draws. Images are
resampling units for Attribute and compact Association. Spatial retains its
48 pair clusters. Prompt variants and conditions stay together within each
unit. Cellwise intervals describe uncertainty within each sampled task.

For the Color example in Section~\ref{sec:cross-task}, LLaVA's
attribute contrast is $+.057$ with a 95\% interval of $[.037,.076]$.
Qwen3's estimate is $-.072$ with $[-.214,.068]$.

The compact association role-permutation comparison covers nine models,
excluding GPT-4.1-mini, which was added to the main panel later.
It uses exact one-sided image-level sign
flips with Benjamini--Hochberg correction within each family
\citep{benjamini1995fdr}. At $q=.05$, 9/9 Holding and 6/9 Attached-to
tested models remain positive by that test. Leave-one-image-out mean
contrasts stay positive in 9/9 and 7/9. These comparisons measure
responses to coherent versus malformed or type-incongruent fields.

\subsection{Spatial prediction and paired inference}

For Table~\ref{tab:cue-ablation}, we add paired precision estimates to
the existing accuracy contrasts after observing the original results.
Each of the 48 image pairs contains 16 question/template units per
condition. We average within each pair and jointly resample pairs across
conditions and models, using 20,000 percentile-bootstrap draws.
The intervals are pointwise 95\% intervals. The direct Qwen3-minus-Qwen2
contrast preserves this pairing.

The prediction comparison asks whether the full cue response tracks support
for separately worded judgments. It fits coefficients on 24 images and
tests without refitting on separate 48- and 96-image sets. We report
\begin{equation}
\dmse=\mathrm{MSE}_{\rm four\ cells}-\mathrm{MSE}_{\rm three\ cells},
\end{equation}
so negative values favor the complete response.

The 24-image training baseline is unregularized ordinary least squares with columns
$[1,z_{00},s_E,s_P]$, where $s_E=z_{10}-z_{00}$ and
$s_P=z_{01}-z_{00}$. These are simple effects relative to the generic
condition. The augmented predictor adds $\gint_{\rm rel}$.
The target is the image-averaged
truth-oriented margin from separately worded direct Yes/No judgments.
The development target symmetrizes predefined truth-polarity presentations.

The External-48 analysis uses 20,000 bootstrap draws.
Qwen3 reports $-1.623$ with interval $[-3.139,-.334]$ and
Qwen2 $+.001$ with $[-.007,.009]$. Spatial-96 reuses the coefficients
without refitting. Its estimates are $-1.222$ with $[-2.049,-.387]$ for
Qwen3 and $-.002$ with $[-.007,.002]$ for Qwen2. For Spatial-96,
all six models share 20,000 bootstrap draws of 48 pairs.

Adding $\gint$ to the three-cell baseline is an invertible linear
reparameterization of the four raw conditions. It therefore tests the
value of the fourth observation. Appendix~\ref{app:prediction-budget}
compares that observation with an ordinary query at the same query budget.

\subsection{Association prediction and token-length sensitivity}

The separate type-valid study uses grouped leave-one-image-out incremental
prediction of independently worded direct criteria. Its convention is
$\mathrm{MSE}_{\rm base}-\mathrm{MSE}_{\rm augmented}$ (positive favors
the augmented model), the reverse of the historical Spatial convention.
The target is correct-answer probability. The baseline uses answer order,
ordinary score level, the two averaged cue effects, and model indicators.
The augmented model adds intended-minus-swapped $\gint$. Both use ridge
penalty $10^{-6}$, an unpenalized intercept, training-fold standardization,
and no prediction clipping. Every held-out image's models stay together.
Intervals use 10,000 image-bootstrap draws of computed cross-fit losses.
They are conditional on these fits, not a bootstrap of the full training
procedure.

The length sensitivity adds signed and absolute intended-minus-swapped
input-token-count differences. At zero difference and equal model weights,
10,000 image-bootstrap refits give $+.090$
$[-.612,.805]$ for Holding and $-.616$ $[-1.510,.299]$ for Attached-to.
All 3,960 cell/template comparisons have matched intended/swapped token
counts, leaving the point estimates unchanged. Token length is balanced
so these prompts cannot isolate its effect.

Secondary fits separate four local models from the one complete API model,
GPT-4.1-mini. Local-only prediction improvements are $-.002941$
$[-.005859,-.000739]$ for Holding and $-.002514$
$[-.005549,-.000345]$ for Attached-to. API-only improvements are $-.030484$
$[-.068944,.001351]$ and $-.009041$ $[-.015145,-.003597]$.
These secondary regressions assess the API and local subsets separately.
The pooled model is fitted to both subsets jointly.

The pooled role contrast equally weights one API and four local models.
Holm correction covers the four primary comparisons, two per family.
All adjusted $p$-values are 1.0, and the prespecified sign-only joint
criterion is not met. The wide role intervals leave the size and
direction of an intended-role effect uncertain.
\section{Ordinary-Query and Nonlinear Predictive Controls}
\label{app:prediction-budget}

This post-hoc analysis compares alternative measurements at a matched
query budget. It uses six-model outputs from 24 training
images and separate test cohorts, External-48 and Spatial-96. The training
data include both factorial and ordinary judgments of the same four
propositions per image. Training and test images are disjoint.

\paragraph{Prediction target and features.}
For each image we average four truth-oriented proposition margins.
Factorial template 1 supplies $r_{00},r_{10},r_{01},r_{11}$, ordinary
template 1 supplies $u_1$, and ordinary template 2 supplies target $u_2$.
The ordinary questions begin \emph{Is the following statement supported
by the image?} and \emph{Based only on the image, is this statement true?}.
Both require Yes or No. We use known truth labels to orient the
margins toward the correct answer. Response strength under the second
ordinary wording is the target.

Five predictors use (i) $r_{00},r_{10},r_{01}$, (ii) all four cells,
(iii) the first three plus $u_1$, (iv) $u_1$ alone, and (v) a quadratic
ridge basis of the first three. The first four are ordinary least squares
with intercepts. For (v), training-set means and population standard deviations
standardize raw features. Zero-variance features are fixed to zero.
The basis contains three linear terms, three squares and three pairwise
products. We minimize mean squared error plus the sum of squared slopes
(fixed penalty 1, unpenalized intercept), with no tuning.

Models are fitted on the 24 training images
for the single prediction direction specified above. Test cohorts are
reported separately, conditional on these fits. We draw 20,000 shared
source-group bootstrap samples: 48 singleton groups for the first test,
48 pairs for the second.
All alternatives use the same resampled groups.

\paragraph{Comparison with ordinary queries.}
Qwen3's four-minus-three differences remain negative:
$-1.924$ $[-3.568,-.601]$ and $-1.355$ $[-2.048,-.695]$.
Against quadratic-three they are $-.687$ $[-1.962,.563]$ and
$-2.349$ $[-4.229,-1.072]$. Ordinary-query alternatives are nevertheless
substantially stronger for this target (Table~\ref{tab:prediction-budget}).
NVIDIA's first-cohort equal-query interval includes zero.
Thus the fourth label condition adds predictive information for Qwen3,
while a separately worded ordinary query is more effective for this target.

\begin{table*}[t]
\centering
\small
\setlength{\tabcolsep}{5pt}
\begin{tabular*}{\linewidth}{@{\extracolsep{\fill}}lrrrrrl@{}}
\toprule
Model & \shortstack{3 label\\conditions} & \shortstack{4 label\\conditions} & \shortstack{3 conditions\\+ ordinary} & \shortstack{Ordinary\\only} & \shortstack{Quadratic\\3 conditions} & \shortstack{4 conditions $-$\\(3 + ordinary) [95\% CI]} \\
\midrule
Feature queries & 12 & 16 & 16 & 4 & 12 & 16 versus 16 \\
\midrule
\multicolumn{7}{l}{\textit{External-48: 48 test images, 48 source groups}} \\
LLaVA & 0.135 & 0.131 & 0.007 & 0.005 & 0.177 & $0.124$ $[0.078,0.177]$ \\
Qwen2 & 0.870 & 0.873 & 0.014 & 0.014 & 1.292 & $0.858$ $[0.576,1.161]$ \\
Qwen3 & 4.652 & 2.728 & 0.193 & 0.135 & 3.415 & $2.535$ $[1.789,3.390]$ \\
Gemma4 & 2.905 & 1.955 & 0.108 & 0.101 & 3.326 & $1.846$ $[1.016,3.050]$ \\
GPT-4.1-mini & 7.657 & 9.047 & 1.061 & 0.841 & 5.988 & $7.986$ $[5.057,11.232]$ \\
NVIDIA Llama & 0.557 & 0.579 & 0.296 & 0.292 & 1.495 & $0.284$ $[-0.038,0.696]$ \\
\midrule
\multicolumn{7}{l}{\textit{Spatial-96: 96 test images, 48 source groups}} \\
LLaVA & 0.267 & 0.257 & 0.007 & 0.005 & 0.310 & $0.249$ $[0.195,0.306]$ \\
Qwen2 & 1.188 & 1.187 & 0.018 & 0.014 & 1.983 & $1.168$ $[0.850,1.517]$ \\
Qwen3 & 3.529 & 2.175 & 0.143 & 0.126 & 4.523 & $2.032$ $[1.382,2.802]$ \\
Gemma4 & 2.892 & 2.465 & 0.106 & 0.109 & 3.991 & $2.359$ $[1.782,2.984]$ \\
GPT-4.1-mini & 10.700 & 9.171 & 1.303 & 1.234 & 9.271 & $7.867$ $[5.127,11.690]$ \\
NVIDIA Llama & 0.355 & 0.330 & 0.118 & 0.118 & 0.526 & $0.212$ $[0.086,0.347]$ \\
\bottomrule
\end{tabular*}
\caption{Predictive controls at matched query budgets.
MSE is in squared log-odds units. Lower is better. Ordinary denotes a
separately worded question. Query counts cover features per image.
The target costs four more queries.
The paired contrast compares the two 16-query predictors.}
\label{tab:prediction-budget}
\end{table*}

\section{Auxiliary-Hint Evaluation}
\label{app:matched}

This study reuses all 48 Spatial pairs. Each pair contains the same proposition with
opposite truths. Within each image, we retain two question paraphrases and
both A/B-to-Yes/No mappings. Six hints cross role-preserving versus swapped
subject/reference assignments with three cyclic orders of the intact
records. A seventh variant omits the hint. The image, original question,
answer mapping, and generation rule are fixed within every matched block.

\paragraph{Prompt template.}
The box shows the Yes-to-A mapping. Bracketed items are placeholders.
\begin{promptbox}{Spatial-96 role-hint judgment}
Read the image and answer the question. An optional role hint describes
the wording of the question, not facts about the image, and may be
inaccurate. Judge the question, not the hint.\par
Question: \textit{[Question]}\\
Role hint: \textit{[Record]}\par
A: Yes\\
B: No\\
Return only A or B.
\end{promptbox}
The question is either \emph{Is it true in the image that [claim]?}
or \emph{Does the image show that [claim]?}
The record uses \texttt{subject=\textit{[first object]};
relation=\textit{[predicate]}; reference=\textit{[second object]}}.
Swapping exchanges the object values, and rotations move whole fields.
The no-hint record is \texttt{(none)}. The other answer mapping reverses
Yes and No without changing the question or hint.

The same design runs with the real image, the existing neutral gray carrier,
and no image. Identical blind requests shared by the two images are sent
once and evaluated against each truth, giving 5,376 unique requests per
model and 8,064 analysis references. These repeated prompts remain nested
inside 48 independent pair clusters. Both-images-correct requires a correct
answer for each image under a matched variant. Deterministic identical
blind answers cannot satisfy it.

The six models are Qwen3-VL-8B, Qwen2-VL-7B, LLaVA-1.6-Mistral-7B,
GPT-4.1-mini-2025-04-14, GPT-4o-2024-11-20, and NVIDIA-hosted
Llama-3.2-11B-Vision. This study scores generated answers. Generation is greedy, with at most
eight output tokens and no requested log probabilities or forced label
bias. The parser accepts a standalone case-insensitive A/B, optionally
followed by a period or closing parenthesis, with surrounding whitespace.
Other returned answers count as incorrect, with invalid-format rates
reported separately. Unavailable responses remain missing.

NVIDIA responses were collected on two dates with the same prompts and
settings. The hosted service does not expose a weight revision.

Let $A_{CK}, A_{CR}, A_{SK}, A_{SR}$ denote accuracies for role-preserving (C)
or swapped (S) content in canonical (K) or rotated (R) order. R averages
the two rotations equally. Canonical and R receive equal weight:
$A_C=(A_{CK}+A_{CR})/2$ and $A_S=(A_{SK}+A_{SR})/2$. The order effect is
$D_F=(A_{CK}+A_{SK}-A_{CR}-A_{SR})/2$. The interaction is
$I=A_{CK}-A_{CR}-A_{SK}+A_{SR}$. These contrasts also apply to
pair-both-correct accuracy.

Inference uses 20,000 bootstrap draws of the 48 pairs, carrying all
variants and models together. Percentile 95\%
intervals accompany a null-centered one-sided bootstrap test with a +1
Monte Carlo correction. Holm correction applies to exactly two primary
Qwen3 real-image hypotheses, $D_C>0$ and $H>0$. Other effects are
descriptive. API/local means equally weight three models.

\subsection{Average order effects}

Table~\ref{tab:hint-order} reports the average order effect $D_F$ and the
content-by-order interaction for all models. Across the three local
models, mean preserving help is $-.07$ pp $[-1.52,1.41]$ and mean
swapped harm is $+2.52$ pp $[.46,4.56]$.

\begin{table}[t]
\centering
\small
\setlength{\tabcolsep}{3pt}
\newcommand{\mci}[3]{\shortstack{$#1$\\{\scriptsize$[#2,#3]$}}}
\begin{tabular*}{\linewidth}{@{\extracolsep{\fill}}lcc@{}}
\toprule
Model & \shortstack{Canonical $-$\\Rotated, $D_F$}
 & \shortstack{Content $\times$\\Order, $I$} \\
\midrule
Qwen3 & \mci{-3.78}{-5.73}{-1.95} & \mci{8.33}{3.91}{13.15} \\[4pt]
Qwen2 & \mci{-1.04}{-2.08}{-.13} & \mci{1.30}{-.65}{3.65} \\[4pt]
LLaVA & \mci{.59}{-1.63}{2.80} & \mci{-.39}{-3.52}{2.73} \\[4pt]
GPT-4.1-mini & \mci{-.52}{-2.54}{1.37} & \mci{-4.43}{-7.94}{-1.30} \\[4pt]
GPT-4o & \mci{-.46}{-2.73}{1.76} & \mci{-1.17}{-5.34}{2.60} \\[4pt]
NVIDIA Llama & \mci{.46}{-1.95}{2.93} & \mci{7.42}{2.34}{12.50} \\[4pt]
\bottomrule
\end{tabular*}
\caption{Secondary order contrasts (pp), with 95\% pair-bootstrap
intervals. $D_F$ averages the order change across hint contents.
$I$ measures how much larger the preserving-over-swapped gap is in
canonical order.}
\label{tab:hint-order}
\end{table}

\subsection{Observed cells and cue effects}

Table~\ref{tab:matched-cells} reports all five real-image cells. Every
returned real-image answer from Qwen3, Qwen2, LLaVA, and the two GPT models is
format-valid, so their accuracy differences reflect answer choices.
NVIDIA's non-A/B responses remain incorrect under the same scoring rule.
Table~\ref{tab:matched-effects} compares each hint with the unhinted
baseline. The two Qwen3 tests are primary, and the remaining
cross-model comparisons are descriptive. Qwen3's Holm-adjusted
$p$-values are .0112 for $D_C>0$ and .739 for $H>0$.

\begin{table*}[t]
\centering
\small
\setlength{\tabcolsep}{6pt}
\begin{tabular*}{\linewidth}{@{\extracolsep{\fill}}lrrrrr@{}}
\toprule
\multicolumn{6}{l}{\textbf{(a) Scene-level accuracy (\%)}} \\
\addlinespace[3pt]
 & \multicolumn{2}{c}{Role-preserving} & \multicolumn{2}{c}{Swapped} & \\
\cmidrule(lr){2-3}\cmidrule(lr){4-5}
Model & Canonical & Rotated & Canonical & Rotated & No hint \\
\midrule
Qwen3 & 80.99 & 80.60 & 71.88 & 79.82 & 81.51 \\
Qwen2 & 84.11 & 84.51 & 83.07 & 84.77 & 85.68 \\
LLaVA & 55.73 & 55.34 & 53.91 & 53.12 & 53.65 \\
GPT-4.1-mini & 80.73 & 83.46 & 82.03 & 80.34 & 79.69 \\
GPT-4o & 67.45 & 68.49 & 66.93 & 66.80 & 65.62 \\
NVIDIA Llama & 45.83 & 41.67 & 39.06 & 42.32 & 24.22 \\
\midrule
\multicolumn{6}{l}{\textbf{(b) Both images in the pair answered correctly (\%)}} \\
\addlinespace[3pt]
Qwen3 & 66.67 & 65.89 & 46.88 & 63.80 & 67.19 \\
Qwen2 & 68.75 & 70.05 & 67.19 & 70.57 & 72.92 \\
LLaVA & 19.27 & 20.83 & 16.67 & 17.71 & 15.62 \\
GPT-4.1-mini & 66.67 & 71.61 & 67.19 & 65.62 & 65.10 \\
GPT-4o & 43.23 & 43.49 & 41.15 & 40.62 & 35.42 \\
NVIDIA Llama & 3.65 & 3.12 & 0.00 & 3.39 & 1.56 \\
\bottomrule
\end{tabular*}
\caption{Condition-level results underlying Table~\ref{tab:matched-effects}.
Panel (a) scores each image. Panel (b) requires correct answers for both
images in a pair. Invalid-format answers remain incorrect.
Rotated equally averages the two rotations.
Each canonical or no-hint cell has 384 scene-level references and each
rotated cell 768, nested in 48 image pairs. Canonical and the mean rotation
receive equal weight in the contrasts.}
\label{tab:matched-cells}
\end{table*}

For Qwen3, the corresponding scene-level accuracy intervals are
$[72.66,88.54]$ for role-preserving/K, $[72.66,87.76]$ for
role-preserving/R, $[64.84,78.39]$ for swapped/K,
$[72.27,86.59]$ for swapped/R, and $[73.70,88.54]$ for no hint
(percent, 95\% pair-bootstrap intervals). Figure~\ref{fig:matched}
plots the condition means relative to no hint. Table~\ref{tab:matched-effects}
gives paired intervals for the order-averaged contrasts.

For Qwen3's pair-both-correct outcome, $D_C=10.94$ pp
$[4.17,18.23]$, $H=-.91$ pp $[-5.34,3.78]$, and $J=11.85$ pp
$[3.65,20.44]$. The order effect is $-8.07$ pp $[-11.85,-4.56]$ and
interaction $17.71$ pp $[9.38,26.82]$. Swapped hints again drive the
relative gap, while preserving hints stay near baseline.

\subsection{Blind inputs, format validity, and missingness}

The five models without real-image format failures give exactly 50\% accuracy and zero
pair-both-correct in every gray-carrier cell. Qwen3, Qwen2, LLaVA, and
GPT-4.1-mini do the same without an image. Because identical blind requests
are shared by opposite-truth images, any valid deterministic A/B response
gets exactly one member correct. The 50\% result follows directly
from the opposite-truth pairing.

GPT-4o's image-free answers are mostly outside the required A/B format:
380/384 references are invalid for role-preserving canonical hints and all
references are invalid in the other four cells. Its corresponding
format-inclusive accuracies are .52\% and 0\%. Thus its no-hint
real-minus-image-free difference of 65.62 pp $[59.64,71.88]$ combines
format compliance with decision performance. The real-minus-gray difference
is 15.62 pp $[9.64,21.88]$. For Qwen3, Qwen2, LLaVA, and GPT-4.1-mini,
no-hint real-minus-either-blind differences are respectively 31.51 pp
$[23.70,38.54]$, 35.68 pp $[30.47,40.62]$, 3.65 pp $[-1.30,8.59]$,
and 29.69 pp $[21.61,36.98]$. Cellwise real-minus-blind contrasts remain
available in the accompanying results. No blind pair has both answers correct.

Blind requests each contribute two scene references.
Table~\ref{tab:nvidia-controls} separates
accuracy from format compliance. Its real-image invalid-format counts are
69/384, 164/768, 91/384, 165/768, and 211/384 in the five table columns.
All requests returned a response.
Invalid answers fall from 54.95\% without a hint to 17.97--23.70\% with
hints. Thus NVIDIA's accuracy gains cannot be attributed solely to better
visual decisions. The three local models' descriptive
mean effects are $D_C=2.45$ pp $[1.00,4.06]$, $H=-.07$ pp
$[-1.52,1.41]$, $J=2.52$ pp $[.46,4.56]$, $D_F=-1.41$ pp
$[-2.45,-.37]$, and $I=3.08$ pp $[1.26,4.95]$. The complete API stratum has
$D_C=1.69$ pp $[.50,2.93]$, $H=8.09$ pp $[6.40,9.83]$, and
$J=-6.40$ pp $[-8.01,-4.77]$, including NVIDIA's format effects.
The stratum means describe the three models in each group.

\begin{table*}[!t]
\centering
\small
\setlength{\tabcolsep}{6pt}
\begin{tabular*}{\linewidth}{@{\extracolsep{\fill}}lccccc@{}}
\toprule
NVIDIA input & Role-preserving / K & Role-preserving / R & Swapped / K & Swapped / R & No hint \\
\midrule
Real image & 45.83 / 17.97 & 41.67 / 21.35 & 39.06 / 23.70 & 42.32 / 21.48 & 24.22 / 54.95 \\
Gray carrier & 46.61 / 6.77 & 47.27 / 5.47 & 47.40 / 5.21 & 47.40 / 5.21 & 45.05 / 9.90 \\
No image & 27.34 / 45.31 & 32.81 / 34.38 & 21.35 / 57.29 & 31.51 / 36.98 & 35.42 / 29.17 \\
\bottomrule
\end{tabular*}
\caption{NVIDIA's accuracy / invalid-format rate, both in percent.
All 5,376 requests have responses. Non-A/B answers count as wrong.
Gray and image-free responses are shared by opposite-truth images.}
\label{tab:nvidia-controls}
\end{table*}

\subsection{Token-count sensitivity}

For each of the five observable tokenizers, all 384 matched real-image
blocks have identical plain-prompt token counts across their six hinted
variants. All 1,152 role-preserving/swapped comparisons have zero count difference.
Local counts use each model's tokenizer, while OpenAI counts use
o200k\_base. This comparison covers plain text. Chat/image overhead
and NVIDIA's server tokenizer are unavailable.
The hinted variants match in token count while varying token content
and position. No-hint prompts are shorter.
\section{Role Hints on External-48}
\label{app:external-hints}

\subsection{Images, prompts, and comparisons}

This experiment uses External-48, the 48-image test set previously
used for prediction (Appendix~\ref{app:prediction-budget}). It is
disjoint from Spatial-96. For each image, we ask two opposing
left/right claims about the same object pair
and both A/B-to-Yes/No mappings. Unlike Spatial-96, which pairs the same claim across two
images, this design pairs opposing claims within an image.

We evaluate Qwen3-VL-8B and GPT-4.1-mini. Three question
presentations cross seven hint conditions. First places the question
before the hint, Last places it after the hint, and Repeat presents it
on both sides. Each presentation includes no hint, three role-preserving
orders, and three swapped orders. This gives 4,032 requests per model
and 192 responses per condition, nested within 48 images.

The box uses the First presentation and Yes-to-A mapping.
Brackets mark placeholders.
\begin{promptbox}{External-48 role-hint judgment}
Read the image and answer the question. An optional role hint describes
the wording of the question, not facts about the image, and may be
inaccurate. Judge the question, not the hint.\par
Question: Is it true in the image that \textit{[claim]}?\\
Role hint: \textit{[record]}\par
A: Yes\\
B: No\\
Return only A or B.
\end{promptbox}
The second mapping reverses Yes and No. Hint records use the fields,
swaps, rotations, and no-hint marker defined in
Appendix~\ref{app:matched}. First and Last contain the same words.
Repeat adds one copy of the question. Generation is greedy, with at
most eight output tokens and the same A/B parser.

\begin{table*}[t]
\centering
\small
\setlength{\tabcolsep}{5pt}
\begin{tabular*}{\linewidth}{@{\extracolsep{\fill}}llrrrrrrr@{}}
\toprule
 & & & \multicolumn{3}{c}{Role-preserving} & \multicolumn{3}{c}{Swapped} \\
\cmidrule(lr){4-6}\cmidrule(l){7-9}
Model & Question & No hint & Canonical & Rotation 1 & Rotation 2
 & Canonical & Rotation 1 & Rotation 2 \\
\midrule
Qwen3 & First & 88.02 & 88.54 & 86.46 & 84.90 & 72.92 & 80.73 & 78.65 \\
 & Last & 88.54 & 85.42 & 77.60 & 75.52 & 69.79 & 79.17 & 76.56 \\
 & Repeat & 89.06 & 89.06 & 88.54 & 84.38 & 72.92 & 83.33 & 81.77 \\
\addlinespace[3pt]
GPT-4.1-mini & First & 85.94 & 88.54 & 85.42 & 86.46 & 80.73 & 85.42 & 84.90 \\
 & Last & 83.85 & 89.58 & 86.98 & 80.73 & 68.75 & 78.65 & 79.17 \\
 & Repeat & 90.10 & 85.94 & 88.54 & 83.85 & 83.33 & 83.85 & 86.98 \\
\bottomrule
\end{tabular*}
\caption{All 21 conditions per model on External-48.
Entries are accuracy (\%). Every returned answer is format-valid.
Each cell covers both truth values and A/B mappings.}
\label{tab:external-all-cells}
\end{table*}

\paragraph{Inference.}
We use 20,000 paired image-bootstrap draws.
All claims, answer mappings, and conditions for an image stay together.
The two prespecified primary Qwen3 comparisons are a positive
First content-by-order interaction and reduced canonical swapped-hint
harm under Last relative to First. Their Holm-adjusted one-sided
bootstrap $p$-values are .0017 and .871. Other intervals are descriptive.
Table~\ref{tab:external-hints} averages the three field orders equally
for help and harm. Spatial-96 instead gives canonical order and the
mean rotation equal weight. Both experiments define the order
interaction as canonical minus mean rotated role gap.

\subsection{Question position and repetition}

Table~\ref{tab:external-position} reports absolute accuracy changes
as well as corrections and newly introduced errors. Moving the question
last does not reduce canonical swapped-hint harm. The estimated harm
reduction is $-3.65$ pp $[-10.42,3.12]$ for Qwen3 and $-9.90$ pp
$[-18.23,-1.04]$ for mini. Repeat yields $-1.04$ pp $[-6.77,4.69]$
and $-1.56$ pp $[-6.77,3.65]$, respectively. These contrasts account
for the no-hint accuracy of each presentation. For example, mini's
swapped accuracy rises by 2.60 pp under Repeat, but its no-hint
accuracy rises by 4.17 pp.

\begin{table*}[t]
\centering
\small
\setlength{\tabcolsep}{4pt}
\begin{tabular*}{\linewidth}{@{\extracolsep{\fill}}llrrr@{}}
\toprule
Model & Change from First & No hint & Preserving, canonical & Swapped, canonical \\
\midrule
Qwen3 & Last & $+0.52\;[-2.08,3.12]$ & $-3.12\;[-9.38,3.12]$
 & $-3.12\;[-9.90,3.65]$ \\
 & Repeat & $+1.04\;[-3.12,5.21]$ & $+0.52\;[-3.12,3.65]$
 & $0.00\;[-4.69,4.69]$ \\
GPT-4.1-mini & Last & $-2.08\;[-5.73,1.04]$ & $+1.04\;[-4.17,6.25]$
 & $-11.98\;[-19.27,-4.69]$ \\
 & Repeat & $+4.17\;[-1.56,9.90]$ & $-2.60\;[-8.33,2.60]$
 & $+2.60\;[-3.12,8.33]$ \\
\bottomrule
\end{tabular*}
\caption{Question-position effects in pp with paired 95\% intervals.
For canonical swapped hints, Last fixes 13 answers and harms 19 for
Qwen3, and fixes 16 and harms 39 for mini. Repeat fixes/harms are 9/9
and 14/9, respectively, out of 192 responses per model.}
\label{tab:external-position}
\end{table*}

\subsection{Paired answer patterns}

This post-hoc analysis examines paired responses. For each image
and answer mapping, the true and
false claims are asked separately with their corresponding hints.
The four patterns are both correct (Yes, No), both reversed (No, Yes),
both No, and both Yes. Table~\ref{tab:external-patterns} shows the
First no-hint and canonical swapped conditions. Each has 96 response
pairs from 48 images and two mappings.

\begin{table}[t]
\centering
\small
\setlength{\tabcolsep}{3pt}
\begin{tabular*}{\linewidth}{@{\extracolsep{\fill}}llrrrr@{}}
\toprule
Model & Hint & Correct & Reversed & No/No & Yes/Yes \\
\midrule
Qwen3 & None & 73 & 0 & 19 & 4 \\
 & Swapped & 44 & 0 & 49 & 3 \\
mini & None & 69 & 0 & 12 & 15 \\
 & Swapped & 61 & 2 & 24 & 9 \\
\bottomrule
\end{tabular*}
\caption{Paired answer counts with the question first.
Canonical swapped hints mainly add No/No pairs.}
\label{tab:external-patterns}
\end{table}

Under canonical swapped hints, 29 Qwen3 pairs move from both correct
to No/No, with none making the reverse transition. Mini has nine such
changes and none in reverse. The predominant change is rejection of
both claims. Semantic answers agree
across A/B mappings on 82 of 96 claims for Qwen3 and 85 for mini.

\paragraph{Cost.}
Qwen3 records 1,207,864 input and
8,064 output tokens. Mini records 1,720,308 input and 4,032 output
tokens, with an estimated API cost of \$0.70. Token accounting differs
across models. GPU monetary costs are unavailable.
\subsection{Field-order weighting}
\label{app:order-weight}

The original Spatial comparison gives canonical order half the weight
and each rotation one quarter. The External-48 comparison weights
the three orders equally. To check whether this difference changes the
interpretation, we apply both conventions to both cohorts for Qwen3 and
GPT-4.1-mini, using the question-first condition on External-48.
This supplementary analysis was specified after the original results.
The order interaction has the same definition in both studies.

We use 20,000 paired bootstrap draws over 48 image pairs for Spatial
and 48 images for External-48. All conditions and answer mappings
within a unit stay together. Table~\ref{tab:order-weight-sensitivity}
reports descriptive intervals for each cohort separately, reusing the
original intervals for rows with the original weighting.

\begin{table*}[t]
\centering\small
\setlength{\tabcolsep}{3pt}
\begin{tabular*}{\textwidth}{@{\extracolsep{\fill}}llccc@{}}
\toprule
Model & Order weighting & Preserving $-$ no hint & Swapped $-$ no hint & Preserving $-$ swapped \\
\midrule
\multicolumn{5}{l}{\textbf{Spatial-96}} \\
Qwen3 & Half canonical & $-0.72\;[-2.99,+1.63]$ & $-5.66\;[-10.03,-1.43]$ & $+4.95\;[+1.43,+8.72]$ \\
Qwen3 & Equal orders & $-0.78\;[-2.95,+1.48]$ & $-4.34\;[-8.25,-0.43]$ & $+3.56\;[+0.43,+6.86]$ \\
GPT-4.1-mini & Half canonical & $+2.41\;[+0.07,+5.08]$ & $+1.50\;[-0.85,+3.84]$ & $+0.91\;[-1.69,+3.71]$ \\
GPT-4.1-mini & Equal orders & $+2.86\;[+0.52,+5.56]$ & $+1.22\;[-1.04,+3.39]$ & $+1.65\;[-0.78,+4.34]$ \\
\midrule
\multicolumn{5}{l}{\textbf{External-48}} \\
Qwen3 & Half canonical & $-0.91\;[-3.26,+1.56]$ & $-11.72\;[-16.15,-7.42]$ & $+10.81\;[+6.90,+14.97]$ \\
Qwen3 & Equal orders & $-1.39\;[-3.99,+1.22]$ & $-10.59\;[-14.76,-6.77]$ & $+9.20\;[+5.56,+13.02]$ \\
GPT-4.1-mini & Half canonical & $+1.30\;[-0.52,+3.39]$ & $-2.99\;[-5.60,-0.26]$ & $+4.30\;[+2.08,+6.77]$ \\
GPT-4.1-mini & Equal orders & $+0.87\;[-1.22,+3.12]$ & $-2.26\;[-4.86,+0.52]$ & $+3.12\;[+1.22,+5.21]$ \\
\bottomrule
\end{tabular*}
\caption{Order-weighting sensitivity in RQ2. Values are pp with descriptive paired 95\% intervals. Half canonical weights the three orders $(1/2,1/4,1/4)$, while Equal orders uses $(1/3,1/3,1/3)$. Original reports use the former for Spatial-96 and the latter for External-48.}
\label{tab:order-weight-sensitivity}
\end{table*}

Changing the weights changes effect sizes but leaves Qwen3's help--harm
interpretation intact in both cohorts. Mini's swapped-hint change remains
positive on Spatial and negative on External-48.

\section{Hint Rewriting and Response Tasks}
\label{app:mitigation}

This section reports all RQ3 comparisons. Hint rewriting changes how
a relation is expressed. The task comparison holds the hint fixed
and replaces claim verification with relation selection.

\begin{table*}[t]
\centering\small
\setlength{\tabcolsep}{4pt}
\begin{tabular*}{\textwidth}{@{\extracolsep{\fill}}lccrr@{}}
\toprule
Hint & Verification, \% [95\% CI] & Selection, \% [95\% CI] & Fixes & New errors \\
\midrule
\multicolumn{5}{l}{\textbf{Qwen3}} \\
No hint & $84.90$ $[74.48, 92.71]$ & $88.54$ $[72.92, 100.00]$ & 16 & 9 \\
A left of B, fields & $85.42$ $[74.48, 93.75]$ & $88.02$ $[76.04, 96.88]$ & 16 & 11 \\
A left of B, sentence & $85.94$ $[76.04, 93.75]$ & $85.94$ $[76.04, 93.75]$ & 16 & 16 \\
B right of A, fields & $63.54$ $[57.81, 69.79]$ & $84.38$ $[72.92, 93.23]$ & 47 & 7 \\
B right of A, sentence & $78.12$ $[69.27, 86.46]$ & $86.98$ $[76.04, 94.79]$ & 35 & 18 \\
B left of A, fields & $71.35$ $[61.98, 80.21]$ & $87.50$ $[75.52, 96.88]$ & 40 & 9 \\
B left of A, sentence & $50.00$ $[50.00, 50.00]$ & $74.48$ $[67.19, 81.25]$ & 48 & 1 \\
A right of B, fields & $51.04$ $[50.00, 52.60]$ & $64.58$ $[57.81, 71.35]$ & 26 & 0 \\
A right of B, sentence & $50.00$ $[50.00, 50.00]$ & $64.06$ $[58.85, 69.27]$ & 27 & 0 \\
\midrule
\multicolumn{5}{l}{\textbf{GPT-4.1-mini}} \\
No hint & $83.33$ $[75.00, 90.62]$ & $84.90$ $[73.44, 93.75]$ & 15 & 12 \\
A left of B, fields & $84.38$ $[76.04, 91.15]$ & $84.90$ $[72.92, 93.75]$ & 10 & 9 \\
A left of B, sentence & $82.81$ $[75.00, 90.10]$ & $86.46$ $[75.52, 94.27]$ & 13 & 6 \\
B right of A, fields & $84.38$ $[75.52, 91.67]$ & $80.21$ $[66.15, 91.15]$ & 19 & 27 \\
B right of A, sentence & $81.25$ $[70.83, 90.10]$ & $79.69$ $[69.27, 87.50]$ & 18 & 21 \\
B left of A, fields & $76.56$ $[68.75, 83.85]$ & $82.29$ $[72.92, 90.10]$ & 20 & 9 \\
B left of A, sentence & $76.56$ $[68.23, 84.38]$ & $75.52$ $[67.71, 82.81]$ & 16 & 18 \\
A right of B, fields & $66.67$ $[59.38, 73.96]$ & $83.85$ $[71.88, 92.71]$ & 55 & 22 \\
A right of B, sentence & $74.48$ $[66.67, 81.77]$ & $78.12$ $[67.19, 87.50]$ & 27 & 20 \\
\bottomrule
\end{tabular*}
\caption{Complete fixed follow-up on 24 images in 12 source pairs. Each cell contains 192 judgments. Intervals resample source pairs, keeping all matched prompts together. Fixes and new errors compare relation selection with claim verification on the same judgments. Hint meanings are illustrated for the claim A left of B.}
\label{tab:mitigation-cells}
\end{table*}

\subsection{Hint Meanings and Prompts}

For the claim A left of B, we use four hints:
\begin{center}
\begin{tabular}{lll}
\toprule
Code & Hint meaning & Relation to claim \\
\midrule
$C$ & A left of B & Same \\
$E$ & B right of A & Same \\
$S$ & B left of A & Opposite \\
$F$ & A right of B & Opposite \\
\bottomrule
\end{tabular}
\end{center}
Thus $C$ and $E$ are equivalent, as are $S$ and $F$. These labels
describe agreement with the claim.
Each expression has a canonical field form and a complete-sentence
form, alongside a ninth, no-hint condition.

Both tasks use the same input below. Object names illustrate the
template. Append exactly one of the two task blocks to this shared
input. Block titles are not part of the prompt.
\begin{promptbox}{Shared input}
Read the image and answer the task. An optional role hint describes
the wording of the claim, not facts about the image, and may be
inaccurate. Use the image to answer.\par
Claim: The cup is to the left of the bowl.\par
Role hint: \texttt{subject=bowl; relation=to the left of; reference=cup}
\end{promptbox}
\begin{promptbox}{Claim verification}
Task: Is the claim true in the image?\par
A: Yes\\
B: No\\
Return only A or B.
\end{promptbox}
\begin{promptbox}{Relation selection}
Task: Where is the cup relative to the bowl in the image?\par
A: The cup is to the left of the bowl.\\
B: The cup is to the right of the bowl.\\
Return only A or B.
\end{promptbox}
The sentence counterpart of this hint is
\texttt{Role hint: The bowl is to the left of the cup.}
The no-hint line is \texttt{Role hint: (none)}.
Both answer mappings are tested for every condition. The original
claim remains visible in relation selection. Task instructions and
answer meanings change together.

\subsection{Cohorts and Evaluation}

The text-only probe uses 16 abstract claim--hint pairs, each with two
answer mappings. It tests explicit agreement between two textual
relations, without an image. Canonical-field accuracy is 19/32 for
Qwen3 and 28/32 for GPT-4.1-mini. Both reach 32/32 with sentence
hints. Other field orders give Qwen3 20/32 and 19/32, and mini 18/32
and 17/32. The field and sentence outputs come from separate batches.

The initial visual rewriting study uses 24 images in 12 source pairs.
The task pilot uses another 12 images in six pairs. A fixed follow-up
then evaluates 24 further images in 12 pairs, retaining the pilot's
complete design. All visual cohorts use the reviewed source-image
inventory described in Appendix~\ref{app:spatial-data}. The follow-up images
and source pairs are disjoint from the earlier RQ3 cohorts.

Each image supplies four claims, crossing object order and relation
direction. Two claims are true and two false. Each uses both answer
mappings. The visual rewriting study has nine conditions, while each
task study crosses those nine with two response tasks. Consequently,
the initial rewriting and task pilot each require 1,728 requests per
model. The fixed follow-up requires 3,456 per model, with 192 judgments
in each task--hint cell. The two tasks have contemporaneous no-hint
baselines, and all conditions are interleaved within matched blocks.

We use Qwen3-VL-8B-Instruct and GPT-4.1-mini-2025-04-14, with
temperature zero, at most eight output tokens, and the image preceding
the text.
Only a complete A or B answer is accepted. Invalid responses count as
errors. There are none in these three visual comparisons.

Eq.~\ref{eq:task-hint} compares the task difference under each hint with
the no-hint task difference. We also report absolute accuracies, corrections,
and new errors to identify what produces the contrast.
The follow-up uses 20,000 paired bootstrap draws over its 12 source
pairs, with all claims, mappings, and conditions kept together.
We report descriptive 95\% intervals.

\begin{table*}[t]
\centering\small
\begin{tabular*}{\textwidth}{@{\extracolsep{\fill}}lcc@{}}
\toprule
Hint & Qwen3, pp [95\% CI] & GPT-4.1-mini, pp [95\% CI] \\
\midrule
A left of B, fields & $-1.04$ $[-6.25, 5.21]$ & $-1.04$ $[-7.29, 4.17]$ \\
A left of B, sentence & $-3.65$ $[-9.90, 3.65]$ & $+2.08$ $[-4.17, 7.81]$ \\
B right of A, fields & $+17.19$ $[10.94, 23.96]$ & $-5.73$ $[-13.02, 1.04]$ \\
B right of A, sentence & $+5.21$ $[1.56, 9.38]$ & $-3.12$ $[-11.46, 4.17]$ \\
B left of A, fields & $+12.50$ $[6.25, 19.27]$ & $+4.17$ $[-4.17, 12.50]$ \\
B left of A, sentence & $+20.83$ $[11.46, 29.69]$ & $-2.60$ $[-11.46, 6.25]$ \\
A right of B, fields & $+9.90$ $[0.52, 19.27]$ & $+15.62$ $[5.21, 25.00]$ \\
A right of B, sentence & $+10.42$ $[2.08, 18.75]$ & $+2.08$ $[-7.81, 10.94]$ \\
\bottomrule
\end{tabular*}
\caption{All eight task-by-hint contrasts in the fixed follow-up. Each subtracts the no-hint task difference from the hinted task difference (Eq.~\ref{eq:task-hint}). Brackets show descriptive paired 95\% intervals.}
\label{tab:mitigation-interactions}
\end{table*}

\subsection{Complete Results and Error Trade-offs}

Table~\ref{tab:mitigation-cells} reports every task--hint cell in the
fixed follow-up. Table~\ref{tab:mitigation-interactions} gives all eight
baseline-adjusted task comparisons. Relation-reversed field hints
($F$) provide the clearest shared recovery across the models. This
pattern appears in both the pilot and the complete follow-up.

In this condition, Qwen3 corrects 26 true-claim judgments and adds no
errors. Mini corrects 55 true-claim judgments but adds four true-claim
and 18 false-claim errors. Counting opposing claims as a pair, the number
answered correctly on both rises from 2 to 28 of 96 for Qwen3 and
from 32 to 69 for mini. Agreement across the two answer mappings falls
from 94/96 to 88/96 for Qwen3 and from 92/96 to 85/96 for mini.

Selection also creates losses for mini. Its accuracy changes by
$-1.04$ pp for $S$ sentences,
$-4.17$ pp for $E$ fields, and $-1.56$ pp for $E$ sentences.
Both models' $F$-field selection accuracies remain below their
unhinted selection accuracies. Relation selection thus recovers part
of the loss caused by these hints.

\begin{table*}[t]
\centering\small
\begin{tabular*}{\textwidth}{@{\extracolsep{\fill}}lrrrrrr@{}}
\toprule
& \multicolumn{2}{c}{Initial rewriting, 24 images} & \multicolumn{2}{c}{Task pilot: Qwen3} & \multicolumn{2}{c}{Task pilot: GPT-4.1-mini} \\
\cmidrule(lr){2-3}\cmidrule(lr){4-5}\cmidrule(lr){6-7}
Hint & Qwen3 & Mini & Verify & Select & Verify & Select \\
\midrule
No hint & 86.46 & 76.04 & 85.42 & 85.42 & 84.38 & 82.29 \\
A left of B, fields & 85.42 & 75.52 & 83.33 & 87.50 & 83.33 & 78.12 \\
A left of B, sentence & 87.50 & 76.56 & 89.58 & 88.54 & 84.38 & 81.25 \\
B right of A, fields & 70.31 & 75.52 & 66.67 & 88.54 & 79.17 & 82.29 \\
B right of A, sentence & 80.73 & 76.04 & 80.21 & 81.25 & 86.46 & 77.08 \\
B left of A, fields & 75.00 & 73.96 & 73.96 & 85.42 & 81.25 & 79.17 \\
B left of A, sentence & 53.12 & 72.40 & 50.00 & 69.79 & 70.83 & 68.75 \\
A right of B, fields & 54.69 & 72.92 & 50.00 & 62.50 & 61.46 & 81.25 \\
A right of B, sentence & 53.65 & 64.58 & 50.00 & 63.54 & 69.79 & 77.08 \\
\bottomrule
\end{tabular*}
\caption{Earlier visual cohorts, shown separately from the fixed follow-up. Entries are accuracy (\%). The initial rewriting comparison has 192 judgments per cell. The task pilot has 96 judgments per cell on 12 images in six source pairs.}
\label{tab:mitigation-earlier}
\end{table*}

Table~\ref{tab:mitigation-earlier} separates the earlier cohorts from
the follow-up. Qwen3's sentence effects recur: $E$ improves by 10.42 pp
initially and 14.58 pp in follow-up verification, while $S$ falls by
21.88 and 21.35 pp. Mini's $F$ sentence effect changes direction,
from $-8.33$ pp initially to $+7.81$ pp in follow-up verification.
In the initial cohort, Qwen3's greater agreement between equivalent
$S$ and $F$ sentence hints also includes more jointly wrong judgments,
rising from 47 to 86 of 192. Here greater consistency accompanies
more shared visual errors.

\paragraph{Query and computational costs.}
Each task condition uses one model call per judgment.
Qwen3 uses 1,169,120 input
tokens and 6,912 output tokens. Mini uses 1,701,808 input tokens and
3,456 output tokens, costing approximately \$0.65.
Relation-selection prompts use longer answer options than verification
prompts, which offer Yes/No.
GPU monetary cost is not estimated. API service fingerprints vary.
Each matched block interleaves conditions.
\subsection{Source-pair influence}
\label{app:existing-sensitivity}

We assess whether individual source pairs dominate the RQ3 follow-up
by omitting each of its 12 pairs once. This supplementary analysis was
specified after the original results. Each omission removes both images
and all their claims, answer mappings, hint conditions, and response tasks.
The remaining cells have 176 judgments from 11 source pairs.
The no-hint task difference is recomputed on those same pairs.

Table~\ref{tab:source-pair-influence} reports all eight task-by-hint
contrasts for both models. Table~\ref{tab:source-pair-transitions} gives
the corresponding corrections and new errors.

\begin{table*}[t]
\centering\small
\setlength{\tabcolsep}{5pt}
\begin{tabular*}{\textwidth}{@{\extracolsep{\fill}}lrrrr@{}}
\toprule
& \multicolumn{2}{c}{Qwen3} & \multicolumn{2}{c}{GPT-4.1-mini} \\
\cmidrule(lr){2-3}\cmidrule(l){4-5}
Hint & Full contrast & Omission range & Full contrast & Omission range \\
\midrule
A left of B, fields & $-1.04$ & $[-3.41,0.00]$ & $-1.04$ & $[-2.27,+1.14]$ \\
A left of B, sentence & $-3.65$ & $[-6.25,-2.27]$ & $+2.08$ & $[+0.57,+4.55]$ \\
B right of A, fields & $+17.19$ & $[+14.77,+18.75]$ & $-5.73$ & $[-7.39,-3.41]$ \\
B right of A, sentence & $+5.21$ & $[+3.98,+5.68]$ & $-3.12$ & $[-5.11,0.00]$ \\
B left of A, fields & $+12.50$ & $[+10.23,+14.20]$ & $+4.17$ & $[+1.70,+6.25]$ \\
B left of A, sentence & $+20.83$ & $[+18.75,+23.30]$ & $-2.60$ & $[-5.11,-0.57]$ \\
A right of B, fields & $+9.90$ & $[+7.95,+11.93]$ & $+15.62$ & $[+13.64,+19.32]$ \\
A right of B, sentence & $+10.42$ & $[+8.52,+12.50]$ & $+2.08$ & $[0.00,+5.68]$ \\
\bottomrule
\end{tabular*}
\caption{Source-pair influence on every baseline-adjusted task contrast in RQ3 (pp). Each omission removes one of the 12 source pairs from both tasks, all hint conditions, and both no-hint baselines. Ranges are minima and maxima over the 12 omissions, not confidence intervals.}
\label{tab:source-pair-influence}
\end{table*}

\begin{table*}[t]
\centering\small
\setlength{\tabcolsep}{5pt}
\begin{tabular*}{\textwidth}{@{\extracolsep{\fill}}lrrrr@{}}
\toprule
& \multicolumn{2}{c}{Qwen3} & \multicolumn{2}{c}{GPT-4.1-mini} \\
\cmidrule(lr){2-3}\cmidrule(l){4-5}
Hint & Fixes & New errors & Fixes & New errors \\
\midrule
A left of B, fields & 11--16 & 6--11 & 6--10 & 6--9 \\
A left of B, sentence & 10--16 & 10--16 & 9--13 & 4--6 \\
B right of A, fields & 40--47 & 4--7 & 16--19 & 20--27 \\
B right of A, sentence & 29--35 & 11--18 & 13--18 & 17--21 \\
B left of A, fields & 34--39 & 6--9 & 16--20 & 7--9 \\
B left of A, sentence & 41--48 & 0--1 & 11--16 & 13--18 \\
A right of B, fields & 21--26 & 0--0 & 47--53 & 15--22 \\
A right of B, sentence & 22--27 & 0--0 & 22--27 & 16--20 \\
\bottomrule
\end{tabular*}
\caption{Ranges of corrections and newly introduced errors over the same source-pair omissions. Each omission retains 176 matched judgments per task--hint cell. Full-cohort counts are reported in Table~\ref{tab:mitigation-cells}.}
\label{tab:source-pair-transitions}
\end{table*}

For relation-reversed field hints, the contrast ranges from 7.95 to
11.93 pp for Qwen3 and from 13.64 to 19.32 pp for mini. No single pair
determines its positive direction. Mini's contrasts for \textit{B right
of A} in fields and \textit{B left of A} as a sentence remain negative
after every omission. The ranges describe the influence of individual
pairs, rather than confidence intervals.

\section{Answer Checking: Matched and Order-Only Comparisons}
\label{app:reliability}

We compare ways to rank the same 96 original answers, one per image in
48 opposite-truth pairs. The matched-variation comparison used in
\S\ref{sec:reliability} varies wording and answer labels in both ordinary
and role-based checks. The earlier order-only comparison instead varies
field order in the role-based checks. We report the matched design first
and use the order-only design as a sensitivity comparison.
Both are post-hoc analyses, with scoring rules specified before ranking.

\begin{table*}[t]
\centering
\small
\setlength{\tabcolsep}{5pt}
\begin{tabular*}{\linewidth}{@{\extracolsep{\fill}}lrrrrl@{}}
\toprule
 & & \multicolumn{3}{c}{Role checks: fixed field order} & \multicolumn{1}{c}{} \\
\cmidrule(lr){3-5}
Model & Ordinary & Canonical & Rotation 1 & Rotation 2 & Mean role $-$ ordinary [95\% CI] \\
\midrule
Qwen3 & \textbf{14.57} & 16.92 & 16.86 & 16.21 & $+2.09\;[-0.30,5.13]$ \\
Qwen2 & 3.54 & 3.53 & \textbf{3.48} & 3.51 & $-0.04\;[-0.48,0.35]$ \\
LLaVA & 44.87 & 44.87 & 46.69 & \textbf{42.99} & $-0.02\;[-3.73,3.54]$ \\
GPT-4.1-mini & 15.88 & 14.20 & 12.38 & \textbf{11.42} & $-3.22\;[-6.20,-0.77]$ \\
GPT-4o & 29.29 & 22.86 & \textbf{22.36} & 26.23 & $-5.47\;[-10.16,-0.57]$ \\
NVIDIA Llama & 67.60 & 62.96 & \textbf{62.90} & 65.85 & $-3.70\;[-8.78,2.04]$ \\
\bottomrule
\end{tabular*}
\caption{Matched-budget answer verification (AURC $\times100$, lower is better).
All strategies use the same initial answer and three checks with matched
wording and answer labels on 96 images. Bold marks row minima.
The last column is mean role AURC minus ordinary AURC, with 95\% intervals
over 48 image pairs. Qwen3 is the primary development comparison.}
\label{tab:matched-reliability}
\end{table*}

\subsection{Fixed answers, scoring, and evaluation}

The original answer $a_0$ uses the real image, no hint, question version
0, and the mapping Yes$\mapsto$A. We keep this answer fixed for evaluation.
All A/B answers are normalized to Yes/No using their public mappings.
Each strategy compares $a_0$ with three auxiliary answers:
\begin{equation}
u(a_0)=\sum_{j=1}^{3}[a_j\ne a_0].
\end{equation}
Brackets indicate disagreement. Larger scores indicate less reliability.
Each strategy costs four queries including $a_0$.
The score uses only disagreement with the initial answer, not ground truth.

We use the same A/B parser. An invalid original answer
counts as an error and receives score 3 in every arm. An invalid auxiliary
answer counts as a disagreement. All six models have complete responses.
The five other
models have no invalid answers in the selected real-image queries.
NVIDIA has 61 invalid original answers among 96, so its ranking results
include detection of format failures rather than visual errors alone.
Their fixed-answer error rates below need not equal one minus the earlier
no-hint accuracies, which average all four wording/mapping combinations.

\paragraph{Error ranking and ties.}

We retain the lowest-scoring answers and evaluate their error rate.
The area under the risk--coverage curve (AURC) averages risk at the
96 nonempty retained counts. Ties receive
their exact expected risk under uniform ordering. For a tied group with
$n$ answers and $e$ errors, preceded by $P$ answers and $C$ errors, its
$t$-th retained position has expected risk $(C+te/n)/(P+t)$.
Neither truth nor input order breaks ties. AUROC gives half credit to
ties and requires both correct and incorrect answers.

Both comparisons use 20,000 paired bootstrap draws of the 48 image
pairs, recomputing metrics and contrasts on each draw. All arms and
models use the same sampled pairs. Intervals are the 2.5th and 97.5th
percentiles. Qwen3's primary comparisons are defined below, and other
model comparisons are descriptive.

\subsection{Matched wording and answer-label changes}
\label{app:matched-reliability}

Table~\ref{tab:matched-reliability} matches the changes in wording and answer labels between
ordinary and role-based checks. This analysis uses the
same outputs and original answers as the
order-only comparison (Appendix~\ref{app:order-only}). For each
role-preserving order $r_0,r_1,r_2$ (canonical, rotation 1, rotation 2), we use the
same auxiliary combinations as ordinary: question 0/Yes$\mapsto$B,
question 1/Yes$\mapsto$A, and question 1/Yes$\mapsto$B.
Each policy uses four queries including the same no-hint $a_0$.
Jointly the four policies need 13 distinct requests per image.
Each auxiliary query is answered independently of the initial response.
We use the same parser and scoring rules as the order-only comparison.

Each policy is evaluated separately. The primary development summary is Qwen3's mean of the
three role-policy AURCs minus ordinary AURC. Each policy produces
its own ranking before these AURCs are averaged.
All 20,000 shared pair-bootstrap draws use this tie rule.

Qwen3's zero-score bins contain 89/83/87 answers and 15/14/14 errors
for the three orders, versus ordinary's 84 answers and 12 errors.
Its mean disadvantage is smaller than the order-only result and its
interval includes zero (Table~\ref{tab:matched-reliability}). See
Table~\ref{tab:reliability-aurc} for the order-only design.
The GPT comparisons favor role cues. Qwen2, LLaVA, and NVIDIA intervals
include zero. NVIDIA's original answers include 61 format failures,
which receive maximum uncertainty in every strategy.
Matching the variations therefore changes which models benefit from
role-based checking.

\subsection{Order-only sensitivity comparison}
\label{app:order-only}

In the earlier comparison, the ordinary arm uses the other three no-hint
combinations of question version 0/1 and Yes$\mapsto$A/B. The
role-preserving arm holds question 0 and Yes$\mapsto$A fixed and uses the
three field orders of the role-preserving hint. The swapped arm does the
same with the swapped hint. Each arm uses four queries, but the dimensions
that vary are not matched across arms. With $a_0$ shared, the three
arms use ten unique queries.

The primary comparison is Qwen3's role-preserving-minus-ordinary AURC.
Negative would favor the
role-related score. Qwen2 has 19,992 two-class AUROC draws. The other
complete models have 20,000. All AURC and paired-difference intervals use
all 20,000 draws. Table~\ref{tab:reliability-aurc} reports both metrics.

\begin{table*}[!t]
\centering
\small
\setlength{\tabcolsep}{3pt}
\newcommand{\rci}[3]{$#1$\;{\scriptsize$[#2,#3]$}}
\begin{tabular*}{\linewidth}{@{\extracolsep{\fill}}lrcccc@{}}
\toprule
\multicolumn{6}{l}{\textbf{(a) AURC $\times100$ and paired differences: lower is better}} \\
\addlinespace[3pt]
Model & Error (\%) & Ordinary & Role-preserving & Swapped & Role $-$ ordinary \\
\midrule
Qwen3 & 18.75 & \rci{14.57}{7.00}{23.12} & \rci{18.29}{10.67}{26.68} & \rci{20.31}{12.24}{28.92} & \rci{3.72}{.42}{7.56} \\
Qwen2 & 8.33 & \rci{3.54}{.38}{8.17} & \rci{4.64}{1.13}{9.29} & \rci{5.57}{1.12}{11.71} & \rci{1.10}{-.91}{3.63} \\
LLaVA & 46.88 & \rci{44.87}{35.13}{54.76} & \rci{42.50}{32.78}{52.58} & \rci{41.99}{31.79}{52.56} & \rci{-2.38}{-7.61}{2.81} \\
GPT-4.1-mini & 16.67 & \rci{15.88}{8.39}{24.49} & \rci{13.91}{6.93}{21.90} & \rci{14.25}{7.18}{22.37} & \rci{-1.97}{-5.26}{1.22} \\
GPT-4o & 30.21 & \rci{29.29}{21.08}{37.35} & \rci{26.18}{18.53}{34.05} & \rci{27.75}{19.70}{35.92} & \rci{-3.10}{-7.24}{1.21} \\
NVIDIA Llama & 80.21 & \rci{67.60}{55.59}{79.01} & \rci{58.79}{49.24}{69.48} & \rci{57.04}{47.40}{67.99} & \rci{-8.81}{-14.68}{-1.95} \\
\bottomrule
\end{tabular*}
\par\medskip
\setlength{\tabcolsep}{10pt}
\begin{tabular*}{\linewidth}{@{\extracolsep{\fill}}lccc@{}}
\toprule
\multicolumn{4}{l}{\textbf{(b) Error-detection AUROC: higher is better}} \\
\addlinespace[3pt]
Model & Ordinary & Role-preserving & Swapped \\
\midrule
Qwen3 & .63 [.52,.76] & .51 [.47,.58] & .45 [.38,.54] \\
Qwen2 & .78 [.63,.94] & .74 [.56,.91] & .68 [.49,.90] \\
LLaVA & .55 [.48,.61] & .58 [.50,.65] & .57 [.49,.64] \\
GPT-4.1-mini & .53 [.43,.64] & .60 [.50,.71] & .58 [.48,.70] \\
GPT-4o & .52 [.42,.61] & .59 [.51,.67] & .56 [.49,.63] \\
NVIDIA Llama & .77 [.67,.87] & .90 [.83,.96] & .91 [.85,.97] \\
\bottomrule
\end{tabular*}
\caption{Order-only sensitivity comparison on 96 fixed original answers
per model. The role arms vary field order. Ordinary checks vary wording
and answer labels. Both panels use the same answers, arms and image pairs.
Brackets are 95\% source-group bootstrap intervals. Panel (a) reports
AURC and its paired difference, multiplied by 100. Error is the original
answer error rate and the expected AURC of random ranking. Panel (b)
reports AUROC. Qwen3's role-preserving-minus-ordinary AURC is the
designated development comparison.}
\label{tab:reliability-aurc}
\label{tab:reliability-auroc}
\end{table*}

\subsection{Errors in Stable Responses}

Qwen3 has 18 wrong original answers. Its role-preserving arm assigns score zero
to 93 answers, including 17 errors. The ordinary arm has 84 zero-score
answers with 12 errors. The swapped arm has 78 with 16 errors. At 50\%
coverage, expected risks are respectively 18.28\%, 14.29\% and 20.51\%
for role-preserving, ordinary and swapped. The large ties mean that a stable
response cannot select among many right and wrong answers. Swapped
consistency has worse secondary AURC than ordinary:
$+5.74$ $[.74,10.89]$ on the $\times100$ scale.

The three local models' descriptive mean AURCs are 21.00, 21.81 and
22.62 (ordinary/role-preserving/swapped), with role-preserving-minus-ordinary
$.81$ $[-1.50,3.15]$.
The complete API stratum has mean AURCs 37.59, 32.96 and 33.01, with
role-preserving-minus-ordinary $-4.63$ $[-6.98,-2.00]$. The API mean
includes NVIDIA's format failures.

\subsection{Query and Computational Costs}

In the order-only comparison, equal query counts do not match token use.
Mean per-answer input-token totals are
1151/1192 (Qwen3), 1428/1469 (Qwen2), 7986/8025 (LLaVA),
1685/1725 (GPT-4.1-mini), and 1522/1563 (GPT-4o),
ordinary/role-preserving, rounded. Swapped matches role-preserving. Tokenizers and
image accounting differ. Mean summed request times are 3.40/3.45/3.42
seconds (GPT-4.1-mini) and 4.28/4.21/4.26 (GPT-4o),
ordinary/role-preserving/swapped. Local timings are unavailable.
Each strategy uses four queries.

\section{Attribute Judgments with Factual Histories}
\label{app:factual}

This exploratory study tests whether earlier mentions of objects and
attributes change a later judgment. It comprises 518 directed
attribute comparisons, 259 base pairs, 257 images, and four models
(Qwen2-VL-7B, InternVL3-8B, LLaVA-1.6-Mistral-7B, SmolVLM-Instruct).
Its InternVL checkpoint differs from the panel's InternVL3.5.

Given annotated assignments $(O_1,A_1)$ and $(O_2,A_2)$, the final
question contrasts $A_1$ with crossed candidate $A_2$ for $O_1$.
FACT histories state \emph{There is a $O$ in this image} and
\emph{There is a $A$ object in this image}. TERM designates these terms
\emph{for the next image question}. Both script assistant \texttt{Okay.}
turns, not measured confirmations. Histories $h_{11}$, $h_{10}$,
$h_{01}$, $h_{00}$ mention $(O_1,A_2)$, $(O_1,A_1)$, $(O_2,A_2)$,
$(O_2,A_1)$, respectively. Under the annotations, FACT existence claims
remain true when the crossed assignment is false.

The word-choice analysis normalizes accepted token mass over two candidate
sets and averages option orders. Against the identical no-history question,
FACT $h_{11}$ increases crossed-candidate support by $+.104$, $+.058$,
$+.084$, $+.073$ (model order above). TERM increases are $+.208$,
$+.236$, $+.306$, $+.137$.

The four-model mean probability-scale interaction is $.0649$
$[.0584,.0714]$ for FACT and $.1455$ $[.1366,.1544]$ for TERM.
FACT-minus-TERM is $-.0805$ $[-.0881,-.0729]$, with image-cluster
bootstrap intervals. Thus TERM produces the larger shift toward the
crossed candidate. TERM also directs attention to the next question,
uses a longer prompt, and mentions the focal attribute more recently.
These differences leave several explanations for the larger effect.
Because the acknowledgment turns are scripted, the study measures the
influence of preceding text rather than the model's confirmation of it.

\end{document}